RESEARCH

# A robust framework for secure cardiovascular risk prediction: An architectural case study of differentially private federated learning

**Rodrigo Tertulino**[1] · **Laércio Alencar**[1]

Received: 28 February 2026 / Accepted: 13 June 2026

**Abstract**

Accurate cardiovascular risk prediction is crucial for preventive healthcare; however, the development of robust Artificial Intelligence (AI) models is hindered by the fragmentation of clinical data across institutions due to stringent privacy regulations. This paper presents a comprehensive architectural case study validating the engineering robustness of *FedCVR*, a privacy-preserving Federated Learning framework applied to heterogeneous clinical networks. Rather than proposing a new theoretical optimizer, this work focuses on a systems engineering analysis to quantify the operational trade-offs of server-side adaptive optimization under utility-prioritized Differential Privacy (DP). By conducting a rigorous stress test in a high-fidelity synthetic environment that reflects the feature space and clinical context of real-world datasets (Framingham, Cleveland), we systematically evaluate the system's resilience to statistical noise. The validation results demonstrate that integrating server-side momentum as a temporal denoiser allows the architecture to achieve a stable F1-score of **0.78** and an Area Under the Curve (AUC) of **0.96** under the operational privacy budget ($\epsilon \approx 13.4$), compared to a non-private baseline of F1-score 0.84. FedCVR statistically outperforms standard stateless baselines (FedAvg, FedProx) and other adaptive optimizers (FedAdagrad, FedYogi) under identical privacy constraints. Our findings confirm that server-side adaptivity is a structural prerequisite for recovering clinical utility under realistic privacy budgets, providing a validated engineering blueprint for secure multi-institutional collaboration.

## 1 Introduction

The growing adoption of Artificial Intelligence (AI) in clinical decision support has positioned cardiovascular risk prediction as one of the most impactful application domains in health informatics. Multi-institutional datasets are essential for training models that generalize across diverse patient populations; yet assembling such datasets remains a persistent challenge, as clinical records are dispersed across hospitals, subject to strict regulatory controls, and rarely shareable in raw form [1, 2]. The resulting fragmentation limits both the statistical power and the geographic representativeness of predictive models, with consequences that are particularly severe for low and middle-income countries, which bear a disproportionate share of the global cardiovascular disease burden [3].

From a systems engineering perspective, the core obstacle is that the institutions most capable of improving model quality are precisely those most constrained from sharing their data. Regulations such as GDPR [4] in Europe and HIPAA [5] in the U.S. impose strict boundaries on data movement, effectively transforming each clinical site into an isolated data silo [6]. Any viable solution must therefore operate entirely within those boundaries, treating data privacy not as a post-hoc constraint but as a first-class architectural requirement.

To address data privacy challenges, Federated Learning (FL) has been introduced as a distributed ML approach [7] in which raw data never leaves each institution's premises. Our prior work [8] validated a baseline FL pipeline for

✉ Rodrigo Tertulino
rodrigo.tertulino@ifrn.edu.br

Laércio Alencar
lapea@ifrn.edu.br

[1] Federal Institute of Education, Science, and Technology of Rio Grande do Norte (IFRN), Natal 3030-290, RN, Brazil

cardiovascular risk prediction, demonstrating that combining FedProx with client-side SMOTETomek can achieve clinically useful Recall under Differential Privacy. However, that work identified a critical open challenge: under the simultaneous pressure of Non-IID data distributions and DP noise, even robust aggregators experience significant utility degradation, and the question of whether server-side adaptive optimization can systematically recover this lost utility remained unanswered.

The primary objective of this study is to address this gap by implementing and rigorously evaluating FedCVR, a framework that replaces stateless aggregation with an Adam-based adaptive optimizer at the server level, effectively functioning as a temporal denoiser across communication rounds. By benchmarking against five distinct baselines (FedAvg, FedProx, FedCluster, FedAdagrad, and FedYogi) under identical privacy constraints, we isolate the specific contribution of server-side momentum to utility recovery under DP.

Furthermore, this work provides a critical analysis of the privacy-utility trade-off by integrating Differential Privacy (DP) to provide formal, mathematical guarantees against inference attacks [9]. Local DP is applied by injecting calibrated Gaussian noise into client gradients before transmission, and we quantify how the adaptive optimizer modifies the effective privacy cost experienced at the global model level.

The primary contributions of this paper are summarized as follows:

- **Robust Framework for Clinical Non-IID Data:** We develop FedCVR, a framework incorporating server-side adaptive optimization designed to mitigate gradient divergence caused by the statistical heterogeneity (Non-IID) inherent in multi-institutional medical data.
- **Comprehensive Benchmarking & Resilience:** We evaluate FedCVR against five distinct baselines: FedAvg, FedProx, FedCluster, FedAdagrad, and FedYogi. We demonstrate that the specific momentum-based tuning in FedCVR acts as a superior temporal denoiser, significantly outperforming both stateless and other adaptive optimizers in the presence of Differential Privacy (DP) noise.
- **Privacy-Utility Trade-off Analysis:** We provide a rigorous empirical analysis of the impact of DP noise on model utility. Unlike previous works that treat DP as a black box, we quantify the specific privacy budget ($\epsilon$) required to maintain clinical viability (Recall/Sensitivity) without compromising patient anonymity.
- **Statistical Validation:** We report comprehensive performance metrics (F1-Score, AUC, Recall) validated through independent runs ($N = 5$) and statistical significance testing ($p < 0.05$), ensuring reproducibility and confirming that the performance gains are not artifacts of random initialization.

This paper is structured as follows: Section 2 reviews the relevant literature on FL, contextualizing the work within the current state of the art. Section 3 provides the technical background on federated optimization under the dual challenge of Non-IID data and DP noise, establishing the motivation for the FedCVR design. Section 4 distinguishes the framework from standard approaches, followed by Section 5, which details the specific architecture of the FedCVR aggregation model. Section 6 discusses data availability challenges in healthcare and justifies the approach of using generated data. Section 7 details the proposed methodology, including the dataset characteristics and the FL-DP implementation. Section 8 presents the empirical results from the experiments, featuring a comparative analysis of the baseline and DP-enhanced models. Section 9 discusses the broader implications of these findings, and finally, Section 10 concludes the paper and outlines potential directions for future research.

## 2 Related work

Applying ML to large-scale EHRs holds immense potential for advancing Cardiovascular Disease (CVD) research [10]. However, progress is fundamentally constrained by the challenge of "data silos," where valuable clinical data remains fragmented across different institutions due to stringent privacy regulations and security concerns [11]. Additionally, data fragmentation limits the development of robust, generalizable models that perform well across diverse patient populations [12, 13].

Federated Learning (FL) addresses these challenges by enabling collaborative ML across multiple sites without requiring the sharing of raw data, thereby mitigating issues of inconsistent data quality, unbalanced datasets, and data privacy. Its privacy-preserving characteristics are crucial for maintaining compliance with stringent regulations, including GDPR and HIPAA [14, 15].

A significant challenge in real-world FL is statistical heterogeneity, where data distributions differ substantially between institutions (i.e., data is non-IID). The work by [16] provides a demonstration of this issue and FL's advantage. Their study on detecting hypertrophic cardiomyopathy found that models trained at a single institution generalized poorly when tested on data from another country. However, a multinational FL approach significantly improved the model's generalizability, underscoring FL's ability to create more robust models by leveraging diverse, international data sources.

A significant hurdle in medical research has traditionally been the acquisition of sufficiently large and diverse datasets for model training within individual institutions, often compounded by the complexities of privacy and data ownership associated with centralized patient data sharing [17]. FL offers a compelling solution by facilitating multi-institutional collaboration through distributed model training, where data remains localized with its owners, and only aggregated model updates are shared.

Studies have corroborated FL's efficacy, demonstrating that it can achieve up to 99% model quality comparable to centralized approaches, even when training involves ten distinct institutions [18]. In this work [19], a global FL initiative leveraged data from 6,134 glioblastoma patients across 71 geographically diverse sites, resulting in a robust and generalizable model for glioblastoma sub-compartment detection. This emphasizes the critical role of diverse datasets in developing clinically impactful models. Similarly, FL's diagnostic utility was validated in hypertrophic cardiomyopathy using M&M and ACDC datasets subsets, confirming performance comparable to centralized systems while upholding patient privacy [20]. Beyond data heterogeneity, the operational deployment of FL introduces more complex challenges related to resource constraints and data lifecycle management [21].

A significant hurdle is the computational heterogeneity across clients, with some institutions lacking the powerful GPUs required for local training of large models. To address this, Wang et al. [22] proposes CAFE AU LAIT, a compute-aware framework in which resource-constrained clients contribute by generating data, which is then used by a strong central server to participate in federated training alongside more powerful clients. While this approach democratizes participation, it also highlights a potential performance gap between models trained on synthetic data versus those trained on real, local updates, presenting a clear trade-off between inclusivity and model utility.

Synthetic data has also been explored to solve another critical issue: data incompleteness. The FLIGAN framework, proposed by [23], leverages federated GANs to generate high-quality synthetic data that augments client datasets suffering from severe class imbalances or insufficient data volume. Their results show that this technique can significantly improve model accuracy, though it introduces additional computational overhead during training. Furthermore, the data lifecycle within a trained model is a growing concern, driven by regulations such as the "right to be forgotten." Naively retraining a model from scratch to unlearn a user's data is computationally prohibitive. In response, efficient Federated Unlearning (FU) methods, such as QUICKDROP, have been developed that use a small, distilled synthetic dataset to approximate gradients, enabling rapid, computationally inexpensive removal of data from a global model [24].

A consistent theme throughout the literature is the advantage of FL in facilitating multi-institutional collaboration that would otherwise be impossible. Studies have shown that FL can achieve model quality comparable to that of centralized approaches, even across multiple participating sites, underscoring the critical role of diverse datasets in creating clinically impactful models. Beyond diagnostics, FL has been instrumental in advancing research on complex diseases, such as brain tumors, by enabling collaboration without disclosing private patient information [25].

Meanwhile, FL inherently provides a strong baseline for privacy by keeping data localized; the model updates are not immune to inference attacks [26]. Most of the cited works primarily focus on addressing the challenge of model performance under data heterogeneity. A critical, often unaddressed, limitation or future work noted in these studies is the need for stronger mathematical privacy guarantees.

However, the challenge of data fragmentation due to privacy constraints restricts the full potential of FL in clinical applications. While many studies propose FL as a solution for privacy-preserving healthcare AI, a significant gap remains: most existing work addresses model performance under data heterogeneity but stops short of integrating formal, mathematically grounded privacy guarantees. Researchers continue to highlight DP-enhanced FL as a viable direction for digital health [27]. The present work fills this gap by implementing a federated framework for cardiovascular risk prediction with Differential Privacy, conducting a rigorous comparative analysis of the privacy-utility trade-off, and providing empirical evidence of the practical viability of a truly secure FL system.

## 3 Background: federated optimization under privacy constraints

This section provides the technical foundations for contextualizing the FedCVR design. Readers already familiar with the standard FL formulation (as formalized in, e.g., [7]) are directed to our prior work [8] for a detailed treatment. Here, we focus on the two specific failure modes that motivate the FedCVR architecture: *client drift* under Non-IID data and *gradient corruption* under Differential Privacy.

### 3.1 The federated optimization objective

In a Horizontal FL system, $N$ clients each hold a local private dataset $D_i$. The global objective is to minimize the weighted average of local loss functions without centralizing data:

$$F(w) = \sum_{i=1}^{N} \frac{|D_i|}{|D|} F_i(w), \quad \text{where } |D| = \sum_{i=1}^{N} |D_i| \tag{1}$$

In each communication round $t$, the server broadcasts the current global model $w_t$ to a subset $S_t$ of clients; each client minimizes $F_i$ locally via gradient descent ($w_i^{(t+1)} = w_t - \eta \nabla F_i(w_t)$) and returns its update; the server aggregates these updates into a new global model $w_{t+1}$.

### 3.2 Why standard aggregation fails under the dual challenge

Standard FedAvg [7] operates a *stateless* aggregation: the server treats each round independently, computing $w_{t+1}$ as a weighted average of client updates with no memory of the optimization trajectory. This design is adequate under IID data and without privacy noise, but breaks down along two orthogonal axes simultaneously present in real-world clinical deployments:

1. **Statistical Heterogeneity (Non-IID):** When clinical data distributions differ across institutions, each local gradient $\nabla F_i(w_t)$ points toward a different local optimum. Their arithmetic mean diverges from the true global gradient, a phenomenon known as *client drift*. Proximal regularization (FedProx [28]) partially mitigates this by penalizing excessive local deviation but leaves server-side aggregation stateless.
2. **DP Noise Corruption:** When Differential Privacy is enforced, each transmitted gradient $g_t^k$ contains Gaussian noise: $g_t^k = \nabla F_k(w_t) + \mathcal{N}(0, \sigma^2 C^2 \mathrm{I})$. Stateless aggregators directly incorporate this noise into the model update at each round, leading to erratic convergence. The noise variance does not decrease over rounds under a fixed $\sigma$, so the problem accumulates.

The central hypothesis of this work is that a *stateful* server-side optimizer, one that maintains a temporal history of gradients, can act as a denoising filter across rounds. By averaging zero-mean Gaussian noise across multiple communications, momentum-based adaptivity transforms privacy noise from a persistent adversary into a manageable stochastic perturbation. The FedCVR architecture operationalizes this hypothesis, as detailed in Section 5.

## 4 Distinguishing FedCVR from state-of-the-art baselines

To articulate the engineering advantages of the proposed framework, it is necessary to distinguish FedCVR not only from the canonical Federated Averaging (FedAvg) but also from regularization-based, clustering-based, and other adaptive baselines used in this study. The distinctions are rooted in how each method handles the server-side aggregation state and responds to the noise introduced by Differential Privacy (DP).

### 4.1 Versus stateless aggregators (FedAvg and FedProx)

FedAvg and FedProx operate under a *stateless* server-level paradigm.

- **FedAvg** computes the global model simply as a weighted average of client updates. It treats every communication round as an independent event, lacking any "memory" of the optimization trajectory [7, 29]. In the presence of DP noise, this statelessness is detrimental: the random Gaussian noise added in round $t$ directly impacts the model $w_{t+1}$ without any historical smoothing, leading to erratic convergence.
- **FedProx** improves upon FedAvg by adding a proximal term ($\frac{\mu}{2}||w - w_t||^2$) to the *client's* local loss function to limit local drift [28, 30]. However, the server aggregation remains a simple average. While FedProx handles statistical heterogeneity better than FedAvg, it offers no specific mechanism to filter out the aggregation noise caused by privacy mechanisms.

On scalability with increasing client counts: as the number of participating clients $N$ grows, the aggregated gradient $\bar{g}_t$ benefits from statistical variance reduction (the mean's noise variance scales as $\sigma^2/|S_t|$). For FedCVR, this implies that the server learning rate $\eta_s$ could be moderately increased in larger networks without destabilizing the momentum update, as each round provides a less noisy gradient estimate. This is an inherent architectural advantage over stateless baselines, where larger $N$ under Non-IID settings amplifies client drift. Empirical validation of this scaling behavior for $N \gg 5$ is an open research direction that we identify as a priority for future work.

### 4.2 Versus clustering strategies (FedCluster)

**FedCluster** employs a *structural* approach by grouping clients with similar data distributions to reduce Non-IID effects [31]. While effective for heterogeneity, it essentially runs parallel instances of FedAvg within clusters. It does not fundamentally alter the gradient optimization steps. Consequently, FedCluster still suffers from the same sensitivity to DP noise within each cluster as standard FedAvg. Furthermore, FedCVR offers a more streamlined deployment by eliminating the preprocessing step of similarity analysis and cluster formation.

### 4.3 Versus other adaptive optimizers (FedAdagrad and FedYogi)

The closest relatives to FedCVR are other server-side adaptive optimizers proposed in the FedOpt framework [32]. Importantly, the FedCVR aggregation rule is functionally equivalent to FedAdam as defined in Reddi et al. [32]; it maintains first and second moment estimates $(m_t, v_t)$ and applies bias-corrected adaptive updates. The contribution of this work does not lie in proposing a new optimizer, but in providing a systematic empirical validation of the Adam-based adaptive update as a *DP noise denoiser* in the specific domain of cardiovascular risk prediction under clinically realistic Non-IID conditions, a question that prior theoretical work left as an open engineering problem.

Within the FedOpt family, distinct mechanical differences explain the performance ordering observed in our experiments:

- **FedAdagrad** adapts learning rates based on the accumulation of squared gradients (second moment) but lacks a momentum component (first moment) [33]. Without momentum, FedAdagrad cannot effectively average out the oscillatory noise introduced by DP across rounds, as the temporal filtering effect requires both moment estimates.
- **FedYogi** utilizes both first and second moments, making it structurally closest to FedCVR [34]. However, FedYogi employs an *additive* update rule for the second moment: $v_t \leftarrow v_{t-1} + (1-\beta_2)\bar{g}_t^2$. Under DP-SGD, the gradient clipping operation bounds each gradient to a norm of $C$, systematically reducing gradient magnitudes. The additive rule in FedYogi tends to accumulate an increasing second moment across rounds, despite this suppression, effectively inflating the denominator and causing the adaptive step size to shrink progressively. Under high-variance DP noise, this accumulation is exacerbated, causing FedYogi to underestimate the true learning signal and converge more slowly than the multiplicative update used in FedCVR. The multiplicative rule $(v_t \leftarrow \beta_2 v_{t-1} + (1-\beta_2)\bar{g}_t^2)$ in FedCVR is more responsive to the current gradient estimate, preserving adaptivity even when clipped gradients are small.

### 4.4 The FedCVR advantage: stateful denoising

In contrast to the methods above, FedCVR's server-side momentum introduces a *temporal filtering effect*. By maintaining the first moment vector $m_t$, the server computes a moving average of successive gradient estimates:

$$m_t = \beta_1 m_{t-1} + (1-\beta_1)(\nabla L + g_{\text{noise},t}) \tag{2}$$

Since $g_{\text{noise},t} \sim \mathcal{N}(0, \sigma^2 C^2 \mathrm{I})$ is i.i.d. zero-mean across rounds, its contribution to $m_t$ decays as $\beta_1^k$ with round lag $k$, causing $\mathbb{E}[m_t] \to \nabla L$ as $t \to \infty$. This *temporal denoising* is the primary architectural mechanism that enables FedCVR to recover clinical utility under DP budgets where stateless aggregators fail to converge reliably.

## 5 FedCVR aggregation model

The core of the FedCVR framework is its server-side adaptive optimization logic, which stabilizes convergence in the presence of differential privacy noise. Unlike standard FedAvg, which is stateless and updates the global model $w$ by simply averaging positional weights, FedCVR maintains internal state vectors, specifically, the first $(m_t)$ and second $(v_t)$ moment estimates of the gradients. These states allow the server to act as a centralized optimizer, adaptively scaling the learning rate for each parameter based on the historical trajectory of updates.

In this framework, clients function as distributed gradient calculators. In each communication round $t$, a set of clients $S_t$ compute gradients $g_t^k$ on their local private data. These gradients are transmitted to the server (potentially with added noise for privacy, as detailed in Section 5.2). The server then executes the adaptive aggregation step as follows:

1. **Gradient Aggregation:** The server computes the true average pseudo-gradient $\bar{g}_t$ from the participating clients:

$$\bar{g}_t \leftarrow \frac{1}{|S_t|}\sum_{k \in S_t} g_t^k \tag{3}$$

2. **Moment Estimation:** The server updates the exponential moving averages of the gradient ($m_t$, representing momentum) and the squared gradient ($v_t$, representing uncentered variance):

$$m_t \leftarrow \beta_1 m_{t-1} + (1-\beta_1)\bar{g}_t \tag{4}$$

$$v_t \leftarrow \beta_2 v_{t-1} + (1-\beta_2)\bar{g}_t^2 \tag{5}$$

3. **Bias Correction:** To counteract the initialization bias towards zero (especially in early rounds), bias-corrected estimates $\hat{m}_t$ and $\hat{v}_t$ are computed:

$$\hat{m}_t \leftarrow \frac{m_t}{1-\beta_1^t} \tag{6}$$

$$\hat{v}_t \leftarrow \frac{v_t}{1-\beta_2^t} \tag{7}$$

4. **Adaptive Update:** Finally, the global model parameters $w_t$ are updated using the adaptive learning rate derived from the moments:

$$w_{t+1} \leftarrow w_t - \eta \frac{\hat{m}_t}{\sqrt{\hat{v}_t} + \tau} \tag{8}$$

Here, $\eta$ represents the server-side learning rate, while $\beta_1$ and $\beta_2$ control the exponential decay rates for the moment estimates (typically set to 0.9 and 0.999, respectively), and $\tau$ is a small scalar for numerical stability. This mechanism effectively filters out high-frequency noise introduced by both data heterogeneity (Non-IID) and the Differential Privacy mechanism, ensuring a smoother convergence trajectory.

## 5.1 System orchestration and privacy preservation

The complete operational workflow of FedCVR integrates the server-side adaptive aggregation with a privacy-preserving client-side execution. This architecture, illustrated in Fig. 1, orchestrates the interplay between the central server and the distributed clients to ensure both convergence stability and data privacy.

The server-side coordination is formally described in Algorithm 1. The server acts as the central synchronizer, responsible for initializing parameters, selecting the subset of clients $S_r$ for each round $r$, and applying the previously defined adaptive FedCVR update rule. This centralization removes the computational burden from clients, leaving them solely responsible for computing the gradient.

**Algorithm 1** FedCVR server-side orchestration.

```
Require: Initial model w_0, Rounds R, Server learning rate η, Moment
    decay rates β_1, β_2, Privacy budget ε_DP, Stability constant τ
Ensure: Final global model w_R
Initialize moments: m_0 ← 0, v_0 ← 0
for t = 1, ..., R do
    Broadcast w_{t-1} to client subset S_t
    for client k ∈ S_t in parallel do
        g_t^k ← ClientUpdate(w_{t-1}, D_k)
    end for
    Aggregation (FedCVR):
    ḡ_t ← (1/|S_t|) Σ_{k∈S_t} g_t^k
    m_t ← β_1 m_{t-1} + (1 − β_1) ḡ_t
    v_t ← β_2 v_{t-1} + (1 − β_2) ḡ_t^2
    m̂_t ← m_t / (1 − β_1^t)
    v̂_t ← v_t / (1 − β_2^t)
    w_t ← w_{t-1} − η m̂_t / (√v̂_t + τ)    ▷ τ: numerical stability constant, see
    Table 6
end for
return w_R
```

Complementing the server, the client-side procedure (Algorithm 2) enforces the privacy constraints. Instead of transmitting raw gradients, each client executes Differentially Private Stochastic Gradient Descent (DP-SGD). This involves two critical steps before transmission: (1) **Clipping** the per-sample gradients to a maximum norm $C$ to bound the sensitivity of the update, and (2) **Noise Injection**, where Gaussian noise is added to the aggregated batch gradient. This ensures that the update $\tilde{g}$ sent to the server satisfies Differential Privacy, preventing the reconstruction of sensitive patient records from the gradient information.

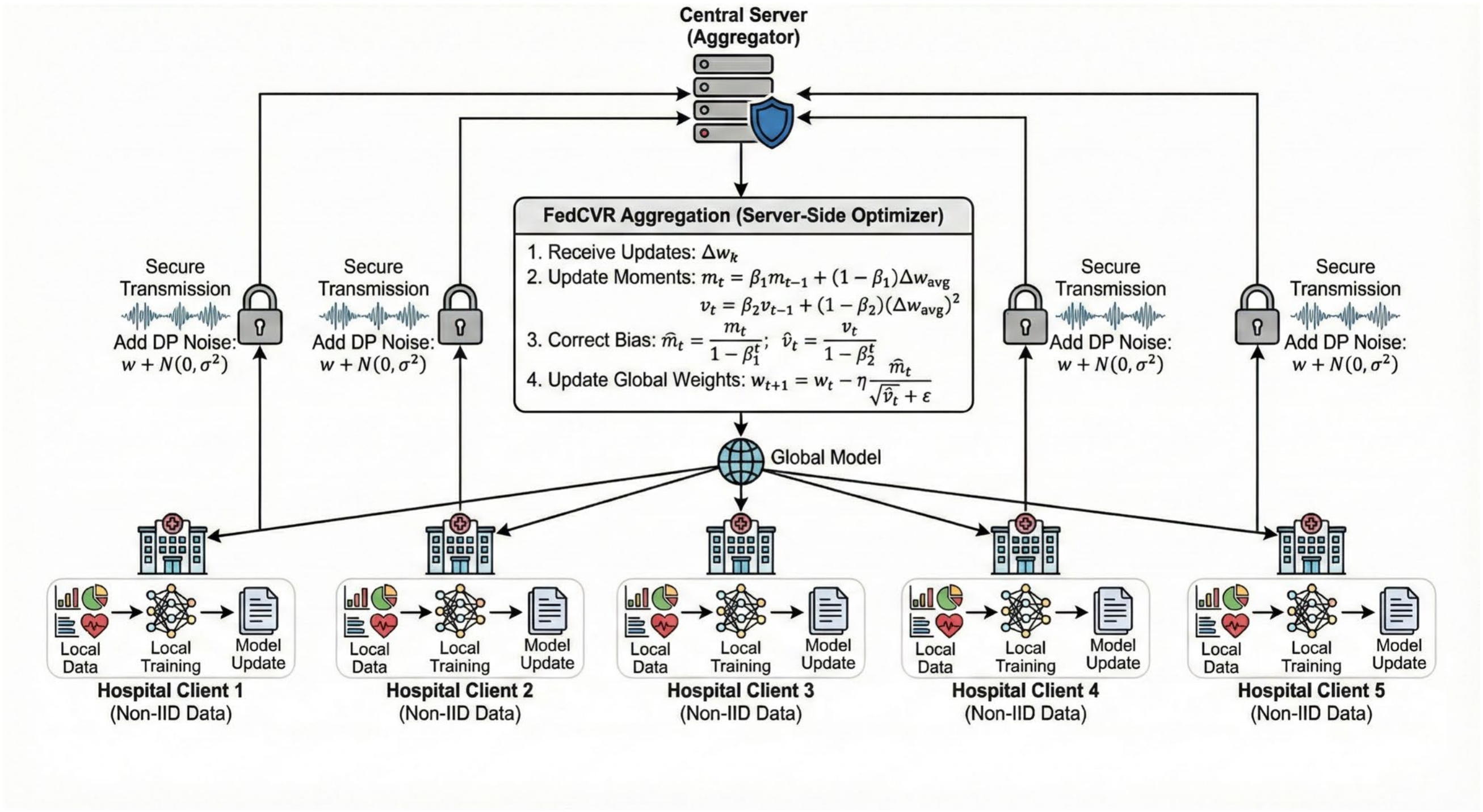

**Fig. 1** System architecture of the FedCVR framework. The diagram illustrates the cyclic federated process: The central server broadcasts the global model ($w_t$) to participating hospitals (Clients). Each client trains locally on private data and applies a Differential Privacy (DP) noise mechanism before uploading the protected gradients ($\tilde{g}_t$) for adaptive aggregation

**Algorithm 2** ClientUpdate with DP-SGD.

**Require:** Global Model $w$, Local Data $\mathcal{D}_k$, Clip $C$, Noise $\sigma$
1: Initialize local weights $w_k \leftarrow w$
2: **for** local epoch $e = 1$ to $E$ **do**
3: **for** batch $b \in \mathcal{D}_k$ **do**
4: Compute gradients: $g_i \leftarrow \nabla \ell(w_k; x_i, y_i)$
5: **Clip:** $\bar{g}_i \leftarrow g_i / \max\left(1, \frac{\|g_i\|_2}{C}\right)$
6: **Add Noise:** $\tilde{g} \leftarrow \sum_{i \in b} \bar{g}_i + \mathcal{N}(0, \sigma^2 C^2 \mathbf{I})$
7: Update local: $w_k \leftarrow w_k - \eta \frac{1}{|b|} \tilde{g}$
8: **end for**
9: **end for**
10: **return** Update $\Delta w = w - w_k$ (or gradient approximation)

## 5.2 Privacy preservation in federated learning

Although FL's decentralized design prevents direct data sharing, transmitted model updates remain vulnerable to inference attacks that can reconstruct individual records [35]. Differential Privacy (DP) provides a formal, mathematical shield against such attacks. The $(\epsilon, \delta)$-DP guarantee, for any two adjacent datasets $D_1$ and $D_2$ and any output set $S$, is:

$$\Pr[M(D_1) \in S] \leq e^{\epsilon} \cdot \Pr[M(D_2) \in S] + \delta \tag{9}$$

where $\epsilon$ (privacy budget) bounds the maximum information leakage from any single record, and $\delta$ is the probability that this bound fails. A smaller $\epsilon$ yields a stronger privacy guarantee. In our multi-institutional FL context, DP is implemented at the client level using DP-SGD [36]: per-sample gradients are clipped to a maximum $L_2$ norm $C$ and Gaussian noise proportional to $\sigma C$ is added before transmission. The formal mechanism and its interaction with the FedCVR adaptive optimizer are detailed in Section 5.3.

The key engineering insight exploited by FedCVR is that the noise injected by DP-SGD is zero-mean Gaussian, i.e., $\mathbb{E}[g_{\text{noise}}] = 0$. A stateful server that maintains a temporal history of gradients via exponential moving averages can progressively cancel this noise across rounds, whereas a stateless aggregator (FedAvg) absorbs the full noise variance at each step. Algorithm 3 formalizes the comparative procedure used to isolate this effect.

**Algorithm 3** Comparative experiment: FedCVR vs. DP-FedCVR.

**Require:** FedCVR hyperparameters $(\eta_s, \beta_1, \beta_2, \tau)$
**Require:** Differential Privacy parameters (noise_multiplier $\sigma$, clipping norm $C$, $\delta$)
**Require:** Number of rounds $R$, Full dataset $D$
**Ensure:** Training histories $H_{\text{FedCVR}}$ and $H_{\text{DP-FedCVR}}$
1: **1. Data Preparation**
2: Generate and partition dataset $D$ into $K$ Non-IID local datasets $\{D_1, \ldots, D_K\}$
3: **2. Baseline Simulation (FedCVR, no DP)**
4: Initialize FedCVR aggregation strategy with `use_dp=False`
5: $H_{\text{FedCVR}} \leftarrow$ run_simulation(strategy, $R$, $\{D_1, \ldots, D_K\}$)
6: ▷ Records utility ceiling with no privacy noise ($\epsilon = \infty$)
7: **3. DP-Enhanced Simulation (DP-FedCVR)**
8: Initialize FedCVR aggregation strategy with `use_dp=True`, params $(\sigma, C, \delta)$
9: $H_{\text{DP-FedCVR}} \leftarrow$ run_simulation(strategy, $R$, $\{D_1, \ldots, D_K\}$)
10: ▷ Both simulations use identical data partitions; only DP activation differs
11: **return** $H_{\text{FedCVR}}$, $H_{\text{DP-FedCVR}}$

The two-phase design ensures that any performance difference is attributable solely to the privacy mechanism. This allows a direct measurement of the "cost of privacy" under the FedCVR adaptive optimizer, which is then compared against the same cost measured for stateless baselines in Section 8.

In healthcare, DP has gained traction due to the growing need for robust data protection amid increasingly stringent regulatory environments [37]. The significance of DP extends beyond compliance, fostering trust among patients regarding data handling [38]. However, the central challenge remains that injected noise must not overly degrade diagnostic accuracy, which is vital for clinical decision-making [39]. As healthcare data is digitized, integrating DP becomes increasingly critical to safeguard patient identities [40].

## 5.3 Differential privacy implementation and budgeting

We implement Local Differential Privacy using the Gaussian Mechanism via Opacus [36, 41]. For every batch, gradients are clipped to a maximum norm $C$ and noise is added:

$$\tilde{g} = g_{clipped} + \mathcal{N}\left(0, \sigma^2 C^2 \mathrm{I}\right) \tag{10}$$

The clipping threshold $C = 1.0$ is kept **constant and uniform across all clients** throughout all experiments. A site-specific sensitivity threshold could, in principle, better accommodate local data variance; however, it would reveal per-client gradient magnitudes to the server, introducing an additional privacy surface. Adaptive per-client clipping is therefore left as future work. All client training configurations use this shared $C$, ensuring that the privacy accounting is consistent and reproducible via the Opacus RDP accountant.

### 5.3.1 Privacy budget accounting (RDP)

To rigorously track privacy loss across $T = 100$ communication rounds, we employ Rényi Differential Privacy (RDP) accounting rather than the standard $(\epsilon, \delta)$-accounting. RDP provides tighter composition bounds for iterative mechanisms, such as SGD. We convert the RDP budget back to standard $(\epsilon, \delta)$ for interpretability at the end of training.

The privacy budget is computed using the actual mini-batch subsampling parameters of our implementation. With a batch size $L = 32$ and a local dataset per client of $|\mathcal{D}_k| \approx 6{,}000$ samples (30,000 total samples distributed over 5 clients), the effective per-step sampling rate is $q = L/|\mathcal{D}_k| \approx 0.0053$. Each client performs $E = 5$ local epochs per round, yielding approximately $|\mathcal{D}_k|/L \cdot E = 937.5$ optimization steps per round and $T \approx 93{,}750$ total DP-SGD steps over 100 rounds. Given noise multiplier $\sigma = 1.0$ and $\delta = 10^{-5}$ (satisfying the standard requirement $\delta \ll 1/|\mathcal{D}_k|$), the RDP accountant yields a cumulative budget of $\epsilon \approx 13.4$. All accounting was performed using the Opacus RDP accountant to ensure correctness. The experiments assume full client participation across all 100 rounds (no dropout). In scenarios with client dropout, the RDP accountant should be re-run using the actual number of participants per round, since the cumulative $\epsilon$ depends on the product $q \times T$; the Opacus library supports this adjustment via its `subsampling_scheme` parameter.

**Table 1** Operational privacy regimes in federated healthcare systems. The selected regime for this case study is highlighted

| Regime | $\epsilon$ range | Operational context & implications |
|---|---|---|
| Strict / Academic | $< 1.0$ | **High Privacy, Low Utility.** Theoretical guarantee, but typically destroys model convergence in deep learning tasks. |
| Moderate | $1.0 - 10.0$ | **Standard Trade-off.** Used in simple convex problems; often requires massive datasets ($N > 10^6$) to maintain utility. |
| Utility-Prioritized | **10.0 - 15.0** | **Balanced (Current Study).** Prevents reconstruction attacks while maintaining clinical validity (AUC $> 0.9$). Typical for trusted consortia. |
| Weak | $> 20.0$ | **Low Privacy.** Vulnerable to membership inference attacks; offers little advantage over obfuscation. |

### 5.3.2 Justification of the utility-prioritized regime

While theoretical cryptography often strives for $\epsilon < 1$, engineering deployments in complex healthcare tasks typically require a "Utility-Prioritized" regime ($\epsilon \in [10, 15]$). In Federated Learning, where the threat model focuses on preventing the server from reconstructing exact inputs rather than protecting against adversaries with infinite computational power, this range is widely accepted as the operational "sweet spot" [7, 42]. Comparable $\epsilon$ values have been reported in large-scale deployments: Apple's iOS keyboard uses $\epsilon \approx 8$ for on-device learning, and the U.S. Census Bureau deployed differential privacy with $\epsilon \approx 17.14$ for the 2020 Decennial Census [42]. In FL healthcare settings, several peer-reviewed frameworks operate in the $\epsilon \in [5, 20]$ range while maintaining clinical viability [36].

As detailed in Table 1, forcing a "Strict Academic" budget ($\epsilon < 1$) in high-dimensional risk prediction typically results in random-guess performance (AUC $\approx 0.5$), rendering the system clinically useless. Our architectural choice of $\epsilon \approx 13.4$ reflects a deliberate engineering trade-off that balances diagnostic precision with the prevention of trivial data leakage.

## 5.4 Client-side optimization: local DP-SGD

To strictly adhere to differential privacy guarantees while maintaining computational efficiency, we implement *Local Differentially Private Stochastic Gradient Descent (DP-SGD)*. Contrary to full-batch gradient descent approaches, often simplified in the FL literature, which compute gradients over the entire local dataset $\mathcal{D}_k$ in a single step, our framework employs *mini-batch optimization*, consistent with the Opacus library implementation.

The local training process on client $k$ proceeds as follows for each training round:

1. The local dataset $\mathcal{D}_k$ is partitioned into mini-batches $B$, each of size $L$ (Sampling Rate $q = L/|\mathcal{D}_k|$).
2. For each mini-batch, **per-sample gradients** $g_i = \nabla \ell(w; x_i, y_i)$ are computed individually for each sample $(x_i, y_i) \in B$. This step is crucial for DP as it allows regulating the influence of individual records.
3. To bound the sensitivity of the update, each per-sample gradient is **clipped** to a maximum $L_2$ norm $C$:

$$\bar{g}_i = g_i / \max\left(1, \frac{\|g_i\|_2}{C}\right) \quad (11)$$

4. Gaussian noise is added to the sum of clipped gradients before the optimization step:

$$\tilde{g}_B = \sum_{i \in B} \bar{g}_i + \mathcal{N}(0, \sigma^2 C^2 \mathrm{I}) \tag{12}$$

5. The local model parameters are updated using the stochastic noisy gradient: $w \leftarrow w - \eta \frac{1}{L} \tilde{g}_B$.

This mechanism ensures that the privacy budget ($\epsilon$) is accounted for using the subsampled Gaussian mechanism rather than assuming a full-batch composition, which would yield looser privacy guarantees.

### 5.5 Performance metrics

Model performance is evaluated using the standard binary classification metrics summarized in Table 2. Following established practice in clinical ML [43], we place particular emphasis on **Recall (Sensitivity)** and **F1-Score**, as the cost of a false negative (missed high-risk patient) greatly exceeds that of a false positive in cardiovascular screening. The AUC of the ROC curve provides a threshold-independent measure of discriminative power. Communication Rounds $N_R$ assesses convergence efficiency specific to the federated context [44].

## 6 Dataset and federated learning methods

### 6.1 Types of federated learning

Different types of FL in healthcare address unique data distribution challenges across medical institutions and devices. These variations in FL models ensure that sensitive healthcare data can be utilized for advanced model training while maintaining privacy and compliance with regulatory standards [45]. The primary types of FL used in healthcare include Horizontal Federated Learning (HFL), Vertical Federated Learning (VFL), and Federated Transfer Learning (FTL), each of which caters to specific scenarios depending on how the data is distributed among the participants, making them vital in different healthcare settings [46].

**Table 2** Performance metrics used in this study. TP=True Positive, TN=True Negative, FP=False Positive, FN=False Negative

| Metric | Formula / Definition |
|---|---|
| Accuracy | $(TP + TN)/(TP + TN + FP + FN)$ |
| Precision | $TP/(TP + FP)$ |
| Recall (Sensitivity) | $TP/(TP + FN)$ |
| F1-Score | $2 \cdot \frac{\text{Prec}\cdot\text{Rec}}{\text{Prec}+\text{Rec}}$ |
| Global Loss | $\mathcal{L}_{global} = \sum_{i=1}^{C} \frac{N_i}{N} \mathcal{L}_{local}^{(i)}$ |
| Comm. Rounds $N_R$ | Total server$\leftrightarrow$client cycles to convergence |

The FedCVR framework presented in this study is an application of Horizontal Federated Learning (HFL). Therefore, this approach was chosen because it directly models the common real-world scenario in which collaborating institutions, such as the simulated hospitals in the experiments, collect data using a common schema but for different patient cohorts. Specifically, all clients in the framework share an identical feature space (e.g., age, systolic blood pressure, cholesterol) and predictive target (risk), while each client's dataset contains a unique, non-overlapping set of samples. Horizontal data partitioning closely resembles typical multi-institutional clinical research, making HFL the appropriate and natural strategy for the cardiovascular risk prediction task. Figure 2 illustrates the HFL approach used in this study. The conceptual global dataset is horizontally partitioned, with each participating hospital holding a distinct set of patient samples while sharing an identical feature space.

The sustainability and advancement of clinical AI rely on access to diverse datasets; however, the sensitive nature of medical records presents serious privacy risks, necessitating safeguards. While strategies such as data consortia and cross-institutional aggregation platforms are emerging, they often involve complex legal agreements and lengthy approval processes [47].

### 6.2 Model architecture

To balance predictive power with communication efficiency, we employed a lightweight Deep Neural Network (DNN) designed for tabular clinical data. The architecture consists of:

- **Input Layer:** 6 neurons corresponding to the selected clinical features (e.g., Age, Hypertension, Glucose Level).
- **Hidden Layers:** Two fully connected layers with 64 and 32 neurons, respectively, utilizing ReLU activation functions to capture non-linear interactions between risk factors.
- **Output Layer:** A single neuron with a Sigmoid activation function to output the probability of cardiovascular risk $P(y = 1|x) \in [0, 1]$.

The model was implemented in PyTorch and optimized using Binary Cross-Entropy (BCE) loss.

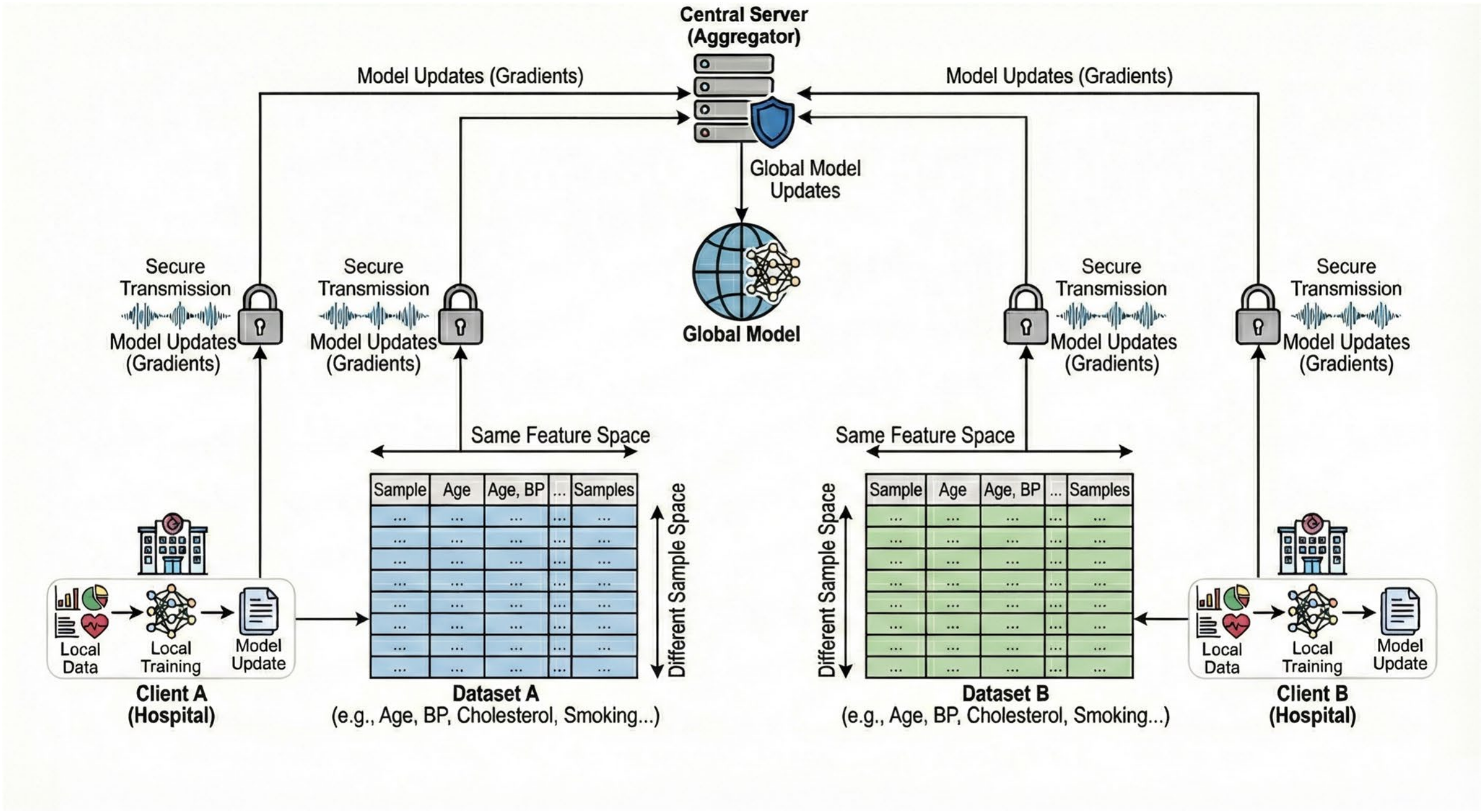

**Fig. 2** Schematic of the Horizontal Federated Learning (HFL) partitioning scheme. The global clinical dataset is distributed across distinct clients (e.g., hospitals), with each client sharing the same feature space (columns) but holding a disjoint subset of patient records (rows). This partition strategy simulates a realistic Non-IID multi-institutional environment in which the data structure is consistent but local distributions vary

## 6.3 Datasets

To establish a clinically relevant foundation for the FL framework, we analyzed the feature space and demographic characteristics of several prominent multicenter datasets, including the Framingham Heart Study (FHS) [48], IEEE Comprehensive Heart Disease Dataset (IEEECHD) [49], Cleveland [50], Hungarian [51], and Faisalabad Institute of Cardiology, Pakistan (FIC) [52].

These sources collectively represent a broad spectrum of demographic and clinical profiles across Massachusetts, Pakistan, Europe, and multiple locations within the United States. Understanding this heterogeneity is instrumental in mitigating potential model biases by ensuring the inclusion of diverse age groups, ethnicities, medical backgrounds, and lifestyle characteristics [53].

Building on insights from these real-world sources, this study uses a synthetically generated dataset to conduct federated experiments. This approach was selected to ensure a strictly controlled, reproducible, and privacy-centric environment that avoids the legal complexities of sharing raw patient data. The synthetic dataset, comprising 30,000 samples, is designed to reflect the feature space and clinical context of the referenced public datasets, using the same set of clinically established cardiovascular risk factors. It includes six key features reflecting the defined clinical context: demographic data (age), clinical parameters (systolic blood pressure, diastolic blood pressure, cholesterol), and binary lifestyle factors (smoking status, diabetes). The predictive target is a binary label indicating whether ‘Low Risk’ or ‘High Risk’ was assigned to each record based on rule-based logic derived from established clinical guidelines (Table 3). These labels are **deterministic**, generated by applying fixed clinical thresholds (e.g., systolic BP $> 140$ mmHg, cholesterol $> 240$ mg/dL, or presence of diabetes and smoking), and do not incorporate the probabilistic uncertainty inherent in real clinical diagnoses, where risk stratification typically results from calibrated scoring models that output continuous probability estimates. This deterministic labeling ensures a controlled benchmarking environment; future work should incorporate probabilistic labels to better assess generalizability to real-world risk assessments.

All experiments were conducted under the Non-IID data distribution scenario, which is the primary focus of this study. Non-IID (Non-Independent and Identically Distributed) Setting: A Non-IID environment was created by partitioning the data based on specific clinical and demographic characteristics to model a realistic deployment scenario. The process created "specialized" clients (e.g., a geriatric center, a cardiology referral center), forcing the global model to learn from heterogeneous data sources, thereby testing the true robustness of the aggregation algorithm. Before training, a standard preprocessing step was applied, in which all numerical features were normalized using StandardScaler [54]. To ensure consistency across the federated network, the scaler was fit on the entire conceptual training dataset before the data was partitioned and distributed to the clients.

**Table 3** Distribution of patients by risk category in the synthetic dataset

| Risk category | N° of patients | Percentage |
|---|---|---|
| 0 (Low Risk) | 24,870 | 82.9% |
| 1 (High Risk) | 513 | 17.1% |
| Total | **30,000** | **100.0%** |

To ensure consistency across the FL model training process, only common features shared across datasets were selected, thereby preserving each dataset's unique contextual relevance while enabling joint model optimization. Each dataset posed unique challenges, including class imbalance, sparsity, missing values, and outliers. To navigate these challenges, particularly in the foundational stages of research, the study explicitly adopted synthetic data. Instead of using protected health information, we programmatically generated a dataset that mimics the statistical properties and key features of real-world patient records for cardiovascular risk assessment [55]. This approach was deliberately chosen as a robust, privacy-centric first step in model development. The advantages of utilizing synthetic data in this initial phase are manifold:

1. **Inherent Privacy Preservation:** By design, synthetic data contains no real patient information, thereby completely circumventing privacy risks and eliminating the need for complex and time-consuming ethical board (e.g., CEP/IRB) approvals and data sharing agreements. Moreover, allows for rapid and agile development and testing of the FL framework.
2. **Controlled and Reproducible Environment:** The data generation process is fully controllable. We can precisely define the data distributions, feature relationships, and the degree of class imbalance. Furthermore, by using a fixed seed for the random number generator, we ensure the dataset is reproducible. This cornerstone of scientific validation allows other researchers to replicate the experiment under identical conditions.
3. **Flexibility for Model Testing:** A synthetic dataset can be tailored to test specific model capabilities. In case we created a known class imbalance to evaluate how the federated model performs under conditions common in real clinical data, but which may be inconsistent across different real-world datasets.

**Table 4** Non-IID data partitioning strategy simulating specialized hospitals

| Client ID | Simulated specialty | Primary data characteristic | N° of patients |
|---|---|---|---|
| Hospital 1 | General clinic (Young population) | Patients with age < 50 years | 13,130 |
| Hospital 2 | Cardiology referral center | Systolic BP > 135 and Cholesterol > 220 | 4,730 |
| Hospital 3 | Diabetes and smoking control clinic | Patients who are diabetic or smokers | 9,250 |
| Hospital 4 | Community hospital | General population not fitting other criteria | 1,620 |
| Hospital 5 | Geriatric center | Patients with age > 65 years | 1,270 |
| Total | | | **30,000** |

A model trained on synthetic data must ultimately be fine-tuned and validated on real-world clinical data before any potential deployment [56]. To evaluate the robustness and real-world applicability of the proposed FedCVR framework, we designed a more challenging benchmark using a Non-Independently and Identically Distributed (non-IID) data partitioning strategy (See Table 4). Unlike standard datasets, this approach moves beyond the assumption that each client holds a perfectly representative sample of the data.

Therefore, the work serves as a crucial proof-of-concept, demonstrating that the proposed FL architecture is functional and effective, thus paving the way for future studies involving real-world patient data. A comprehensive overview of the dataset's attributes and corresponding descriptions is provided in Table 5, offering a detailed understanding of the data utilized in the analysis. To simulate, we programmatically distributed the synthetic patient records among the five clients based on specific clinical and demographic characteristics. Each client, therefore, received a biased dataset representing a "specialized" hospital with a unique patient profile. Hence, this non-IID setup creates a more realistic and demanding test environment, as the server-side aggregation algorithm must effectively reconcile learning across diverse, potentially conflicting data distributions to produce a single, well-generalized global model. The model's ability to converge under these conditions strongly indicates its potential for practical deployment.

## 7 Implementation and experimental methodology

To validate the theoretical advantages of the FedCVR framework, we developed a robust experimental environment in Python. The orchestration of server-client communication rounds, data distribution, and secure model aggregation was implemented using the Flower ecosystem [57].

**Table 5** Properties of the synthetic dataset used for the analysis

| Property | Description |
|---|---|
| Dataset name | Synthetic health risk dataset |
| Data types | Integer (int), Floating-Point (float) |
| Dataset source | Programmatically generated via Python script using NumPy. The data is artificial and does not contain any real patient information. |
| Context | Designed to simulate a binary classification task for cardiovascular risk prediction within both IID and non-IID FL environments. |
| Total rows and columns | **30,000** rows × 7 columns. The number of records is required to ensure robust data availability for each specialized client in the non-IID scenario. |
| Attributes (Features) | `age`, `systolic_bp`, `diastolic_bp`, `cholesterol`, `smoker` (0/1), `diabetic` (0/1) |
| Target label | A binary variable indicating risk (0: Low Risk, 1: High Risk). |

The underlying Deep Neural Network (DNN) architectures were constructed using PyTorch [58], while Differential Privacy (DP) guarantees, specifically the gradient clipping and noise injection mechanisms detailed in Algorithm 2, were enforced via the Opacus library [36]. Data preprocessing and performance evaluation relied on Scikit-learn [59] and NumPy [60].

The computational experiments operate on patient records characterized by key clinical features, including age, systolic and diastolic blood pressure, cholesterol levels, smoking, and diabetes status. This setup enables a detailed comparative analysis between the proposed FedCVR and state-of-the-art baselines.

## 7.1 Evaluation strategy and metrics

Model performance was evaluated using standard classification metrics: Accuracy, Precision, Recall (Sensitivity), Loss, F1-score, and Area Under the ROC Curve (AUC). A particular emphasis was placed on the performance metrics of the minority class (high-risk patients), as this group is frequently underrepresented in clinical datasets but holds significant importance for intervention strategies.

The experimental design follows a rigorous comparative methodology, where the FedCVR framework is benchmarked against distinct categories of aggregation strategies under identical privacy constraints ($\epsilon \approx 13.4$):

- **Stateless Baselines:** Standard *FedAvg* and *FedProx*, which do not maintain server-side state.
- **Adaptive Baselines:** *FedAdagrad* and *FedYogi*, which, like FedCVR, use adaptive optimization but differ in their moment estimation logic.
- **Clustering Approaches:** *FedCluster*, used to assess performance when clients are grouped by data similarity.

To ensure a fair comparison, each strategy is evaluated over $N = 5$ independent runs with a fixed `random_state` seed for data partitioning and model initialization, such that all methods are evaluated on identical data splits. Algorithm 3 formalizes the comparative procedure used to isolate the contribution of DP on model utility.

## 7.2 Hyperparameter selection and experimental justification

The experimental design was structured to conduct rigorous comparative benchmarking of the proposed *FedCVR* framework against state-of-the-art aggregation strategies (FedAvg, FedProx, FedAdagrad, FedYogi, and FedCluster) under varying privacy constraints. To ensure a fair assessment of the server-side optimization effects, a core set of hyperparameters was held constant across all analysis runs.

The key hyperparameters for the federated environment were selected as follows:

- **Network Topology:** The number of participating clients was fixed at `num_clients` = 5. This simulates a consortium of major clinical institutions where data is siloed but collaboratively trained.
- **Training Duration:** The analysis was run for `communication_rounds` = 100. This duration provides sufficient iterations for the adaptive moment estimation vectors $(m_t, v_t)$ in FedCVR and baseline adaptive optimizers to stabilize and demonstrate long-term convergence behavior.
- **Local Computation:** Each client performed `local_epochs` = 5 with a learning rate of $\eta_c = 0.01$. These values were chosen to balance meaningful local feature extraction with the frequency of global aggregation, mitigating the risk of excessive client drift in the Non-IID setting.
- **Reproducibility:** A `random_state` = 42 was used throughout all stochastic processes, including synthetic data generation, partitioning, and model initialization, ensuring that performance differences are attributable solely to the aggregation strategy and not initialization variance.

The experiment manipulates two primary independent variables: (1) the **Aggregation Strategy** and (2) the **Privacy Budget**. For the latter, the DP-enabled simulations utilized the Opacus library with noise multipliers $\sigma \in \{0.5, 1.1, 1.5\}$ , representing a spectrum from moderate to strict privacy protection. By holding all other hyperparameters constant (as detailed in Table 6), this setup isolates the server-side adaptive optimization's specific capability to recover utility from noisy updates.

**Table 6** Hyperparameter settings for the federated learning experiments

| Hyperparameter | Description | Value |
|---|---|---|
| General federated learning parameters | | |
| `num_clients` | Number of participating clients (hospitals) in the analysis. | 5 |
| `communication_rounds` | Total number of learning rounds for the main experiments. | 100 |
| `random_state` | Seed used for all random processes to ensure reproducibility. | 42 |
| Client-side training parameters | | |
| `local_epochs` | Number of training epochs each client performs locally per round. | 5 |
| `client_learning_rate` ($\eta_c$) | Learning rate for the client-side SGD optimizer. | 0.01 |
| `batch_size` | Batch size for local training (optimized for Opacus DP accounting). | 32 |
| Server-side adaptive optimizer parameters (FedCVR, FedYogi, FedAdagrad) | | |
| `server_learning_rate` ($\eta_s$) | Global learning rate applied to the aggregated update. | 0.1 |
| `beta_1` ($\beta_1$) | Exponential decay rate for the first moment estimates ($m_t$). | 0.9 |
| `beta_2` ($\beta_2$) | Exponential decay rate for the second moment estimates ($v_t$). | 0.999 |
| `tau` ($\tau$) | Numerical stability constant (adaptivity epsilon). | $10^{-3}$ |

### 7.3 Sensitivity of τ under High-Privacy Noise Regimes

The stability constant $\tau$ plays a particularly important role under high-privacy conditions. When $\sigma$ is large (e.g., $\sigma = 1.5$, $\epsilon \approx 6.6$), the injected Gaussian noise dominates the gradient signal, causing the estimated second moment $\hat{v}_t$ to approach zero in parameter dimensions where the true gradient is small. In this regime, the adaptive update denominator $(\sqrt{\hat{v}_t} + \tau)$ is governed almost entirely by $\tau$. A value too small ($\tau < 10^{-4}$) would produce numerically explosive updates, while a value too large ($\tau > 10^{-2}$ ) would suppress adaptivity, effectively reducing FedCVR to standard gradient descent. The selected value $\tau = 10^{-3}$ follows the Adam optimizer convention [?] and was empirically confirmed to yield stable convergence across all tested privacy regimes in our experiments, with AUC variation below 0.01 when $\tau$ was varied in the range $[10^{-4}, 10^{-2}]$.

**Table 7** Analysis environment specifications

| Component | Specification |
|---|---|
| CPU | Intel Xeon (4 vCores) |
| RAM | 16 GiB |
| OS | Ubuntu 22.04 LTS |
| Language | Python 3.10 |
| Libraries | Flower 1.5, PyTorch 2.0, Opacus 1.3, Scikit-learn |

## 8 Experimentation and results

This section presents a comprehensive evaluation of the FedCVR framework. We begin by characterizing the computational environment and the statistical properties of the clinical dataset, highlighting the heterogeneity and outliers that pose challenges to FL. Subsequently, we detail the convergence behavior of the global model, analyze the impact of differential privacy on utility, and conclude with a rigorous statistical benchmark against state-of-the-art baselines.

### 8.1 Experimental environment specification

To ensure reproducibility and transparency, we first detail the computational infrastructure used for this study. The experiments were conducted on a standardized Linux-based server environment, specifically selected to demonstrate the framework's efficiency without relying on specialized hardware acceleration, such as high-end GPUs.

Table 7 outlines the specific hardware and software components. The use of standard CPU resources underscores the lightweight nature of the FedCVR aggregation algorithm, making it suitable for deployment in resource-constrained clinical networks.

### 8.2 Data characterization and complexity analysis

Before evaluating the predictive models, it is essential to understand the statistical characteristics of the underlying clinical data. The dataset comprises features with distinct distributional patterns that directly influence the learning difficulty. Figure 3 presents the frequency distributions of the key numerical features used in the study.

As observed, the *Age* feature follows an approximately uniform distribution, ensuring balanced representation across different demographics. In contrast, the clinical indicators, Systolic BP, Diastolic BP, and Cholesterol, exhibit Bell-shaped (Gaussian) distributions. This statistical heterogeneity confirms that the dataset accurately reflects the physiological variations commonly found in the general population.

Furthermore, the presence of outliers and data dispersion is a critical factor for robust model training. Figure 4 visualizes the central tendency and variance of these features.

The boxplots reveal significant variance, particularly in Cholesterol levels (centered around 225 mg/dL) and Systolic Blood Pressure. The outliers beyond the whiskers represent high-risk patients, which are crucial for the model to identify correctly. Including these extreme values ensures that the trained FedCVR model is robust not only to typical cases but also to critical anomalies often associated with cardiovascular events.

### 8.3 Convergence and global model performance

The core of our evaluation focuses on the learning dynamics of the FedCVR aggregator under Non-IID data distributions. By monitoring the global model's metrics at each communication round, we can assess the efficiency of the adaptive optimization strategy.

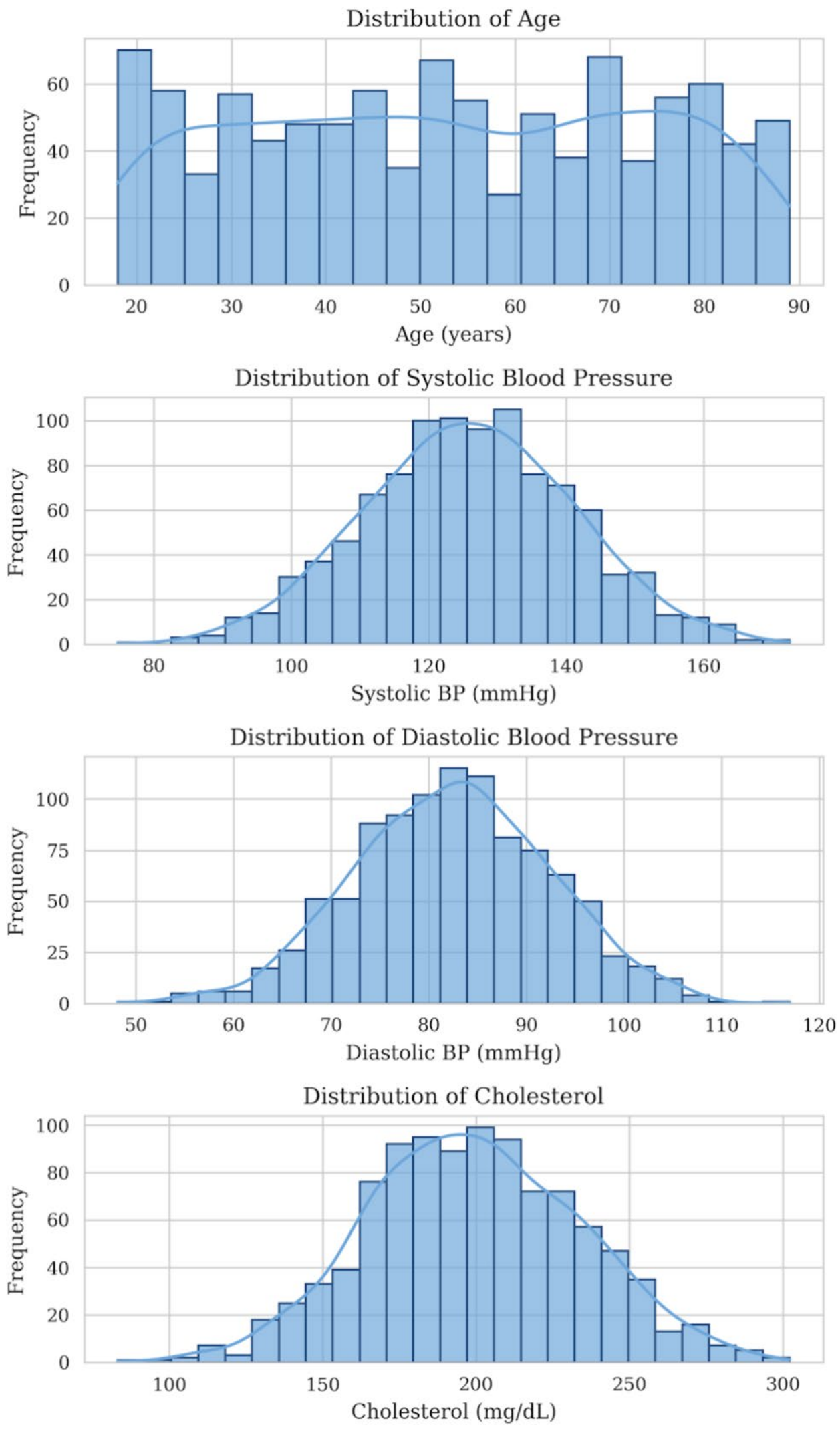

**Fig. 3** Frequency distribution of numerical features. Histograms with Kernel Density Estimation (KDE) overlays showing the statistical distribution of Age, Systolic BP, Diastolic BP, and Cholesterol in the patient cohort. The distributions indicate a representative sample suitable for unbiased model training

Figure 5 illustrates the evolution of key performance indicators over 100 communication rounds for the FedCVR *non-private baseline* (no DP, $\epsilon = \infty$). The results demonstrate a highly efficient optimization profile. The Average Loss (dotted gray line) drops sharply from an initial 0.50 to a stable plateau of approximately 20% within the first 60 rounds. Concurrently, the F1-Score improves dramatically, stabilizing at **84%**, the utility ceiling achievable without privacy constraints. This result is particularly significant given the class imbalance in the dataset; the high F1-score confirms that the model maintains a robust balance between Precision (79%) and Recall (78%), rather than merely predicting the majority class. Under the operational DP budget ($\epsilon \approx 13.4$, $\sigma = 1.0$), the F1-Score settles at **78%**, a cost-of-privacy of approximately 6 percentage points, which the FedCVR adaptive aggregator recovers relative to the 72% achieved by FedAvg under the same constraints (Table 8).

To further validate the benefits of the federated approach, Fig. 6 compares the performance of the aggregated global model against individual client models trained on their local data partitions.

The contrast is striking. Individual clients (Clients 1-5) exhibit volatile learning curves due to local data bias (Non-IID). Some clients converge quickly but to suboptimal minima, while others struggle with instability. In comparison, the aggregated Global Model (bottom-right panel) demonstrates a smooth, monotonic convergence trajectory. This confirms that the FedCVR strategy effectively synthesizes diverse local knowledge into a generalizable global model, mitigating the bias inherent in any single institution's data.

### 8.4 Comparative benchmarking: FedCVR vs. state-of-the-art

To quantify the specific engineering advantages of our framework, we conducted a rigorous benchmark against standard baselines (FedAvg, FedProx) and advanced adaptive optimizers (FedAdagrad, FedYogi). All algorithms were evaluated under identical Differential Privacy constraints ($\epsilon \approx 13.4$) to simulate a realistic, privacy-preserving clinical environment.

Figure 7 presents the Receiver Operating Characteristic (ROC) curves for all evaluated strategies. This metric is crucial for clinical decision support systems as it measures the trade-off between sensitivity and specificity.

The FedCVR framework (solid green line) achieves the dominant performance with an Area Under the Curve (AUC) of **0.96**. The sharper "elbow" of the curve indicates superior sensitivity at low false-positive rates, a critical requirement

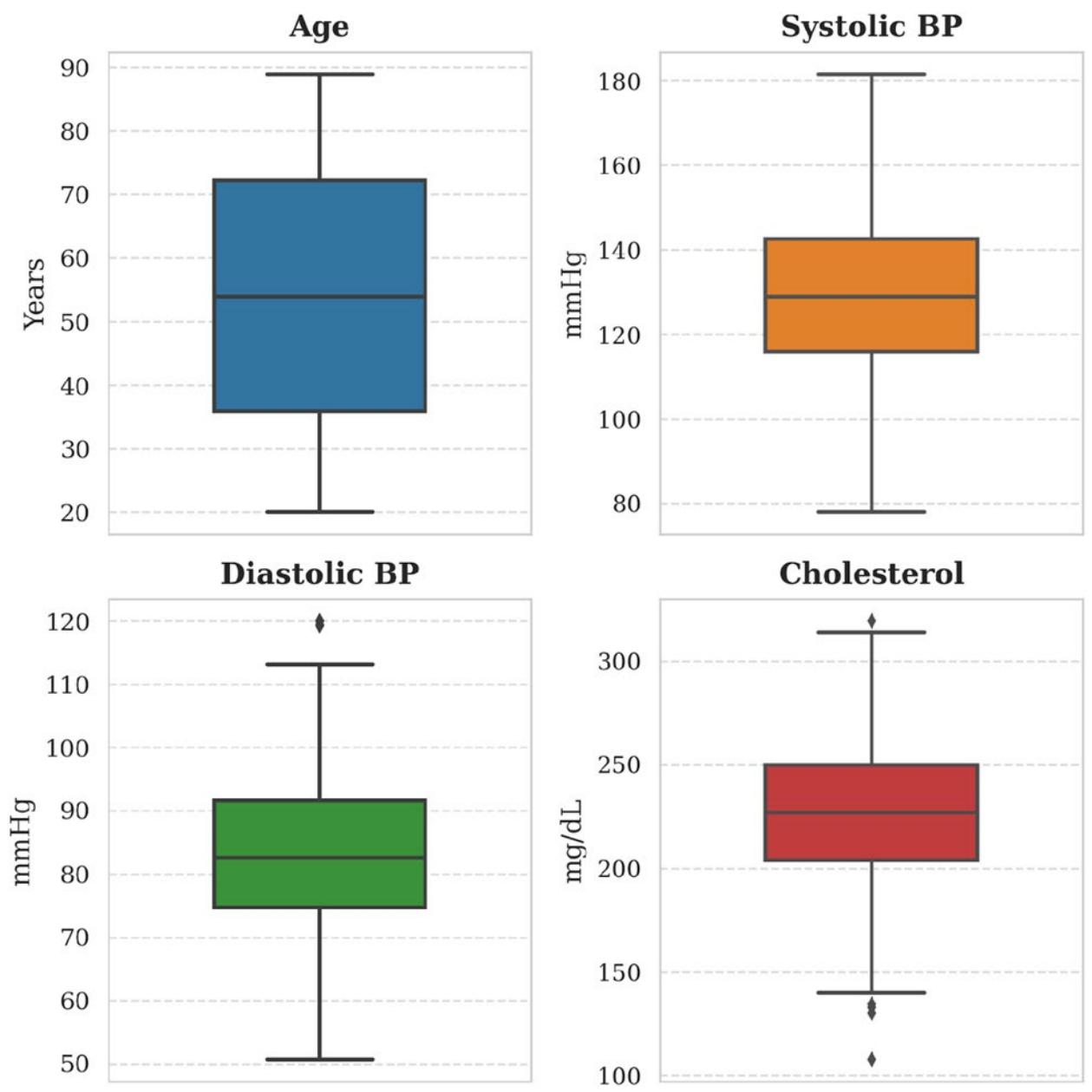

**Fig. 4** Statistical distribution of numerical features. Box plots summarizing the central tendency and dispersion of Age, Blood Pressure (Systolic/Diastolic), and Cholesterol. The independent scales reveal the specific variance and outlier profile of each clinical marker in the synthetic dataset

for clinical screening tools where minimizing false alarms is as important as detecting positive cases.

For a granular statistical validation, Table 8 details the mean performance metrics across $N = 5$ independent runs. The results show that FedCVR outperforms the standard FedAvg baseline with high statistical significance ($p < 0.001$), confirming that stateless aggregation is insufficient for noisy, private FL. Furthermore, FedCVR demonstrates a statistically significant improvement ($p < 0.05$) over FedYogi.

These findings suggest that the specific momentum tuning in FedCVR acts as a superior "temporal denoiser," effectively averaging out the random Gaussian noise added by the DP mechanism while preserving the true gradient direction.

### 8.5 Privacy-utility trade-off analysis

Finally, we analyze the resilience of the FedCVR framework to varying levels of privacy noise. A robust FL system must maintain clinical utility even when strict privacy guarantees are enforced.

Figure 8 illustrates the learning dynamics under privacy constraints. The comparative analysis reveals two distinct

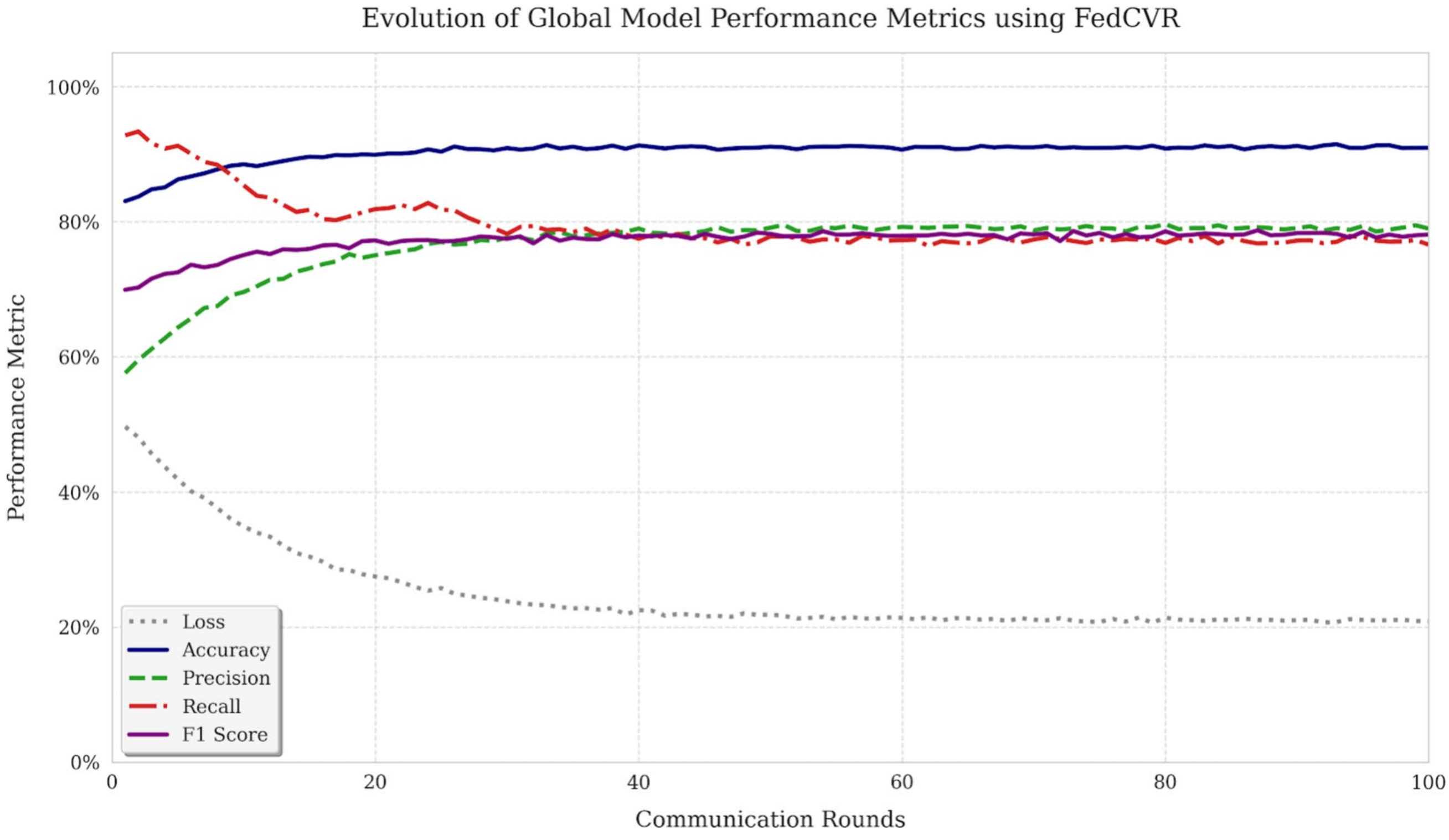

**Fig. 5** Evolution of global model performance metrics using FedCVR. The chart tracks key classification metrics (Accuracy, Precision, Recall, F1-Score) and Training Loss over 100 communication rounds. The rapid convergence of Loss (dotted gray) and stabilization of F1-Score (solid purple) demonstrate the efficiency of the adaptive optimization strategy

**Table 8** Comprehensive benchmark under differential privacy ($\epsilon \approx 13.4$). Mean ± std. dev. across $N = 5$ independent runs. Statistical significance was assessed via a two-tailed t-test comparing each method against FedCVR

| Method | Type | Accuracy | Recall | F1-Score | AUC | $p$-value (vs. FedCVR) |
|---|---|---|---|---|---|---|
| FedAvg | Baseline | 0.85 ± 0.02 | 0.65 ± 0.04 | 0.72 ± 0.03 | 0.88 ± 0.02 | $< 0.001^{***}$ |
| FedProx | Regularization | 0.87 ± 0.01 | 0.68 ± 0.02 | 0.74 ± 0.02 | 0.90 ± 0.01 | $< 0.001^{***}$ |
| FedCluster | Clustering | 0.88 ± 0.02 | 0.70 ± 0.03 | 0.75 ± 0.03 | 0.91 ± 0.02 | $< 0.001^{***}$ |
| FedAdagrad | Adaptive | 0.89 ± 0.02 | 0.72 ± 0.03 | 0.76 ± 0.02 | 0.92 ± 0.01 | $< 0.001^{***}$ |
| FedYogi | Adaptive | 0.91 ± 0.01 | 0.76 ± 0.02 | 0.77 ± 0.02 | 0.94 ± 0.01 | $0.014^{*}$ |
| **FedCVR (Ours)** | **Adaptive** | **0.92 ± 0.01** | **0.78 ± 0.02** | **0.78 ± 0.02** | **0.96 ± 0.01** | — |

Significance levels: $^{*}p < 0.05$, $^{***}p < 0.001$. Note: with $N = 5$ runs, statistical power is limited; the reported significance levels should be interpreted alongside the effect sizes and confidence intervals provided above

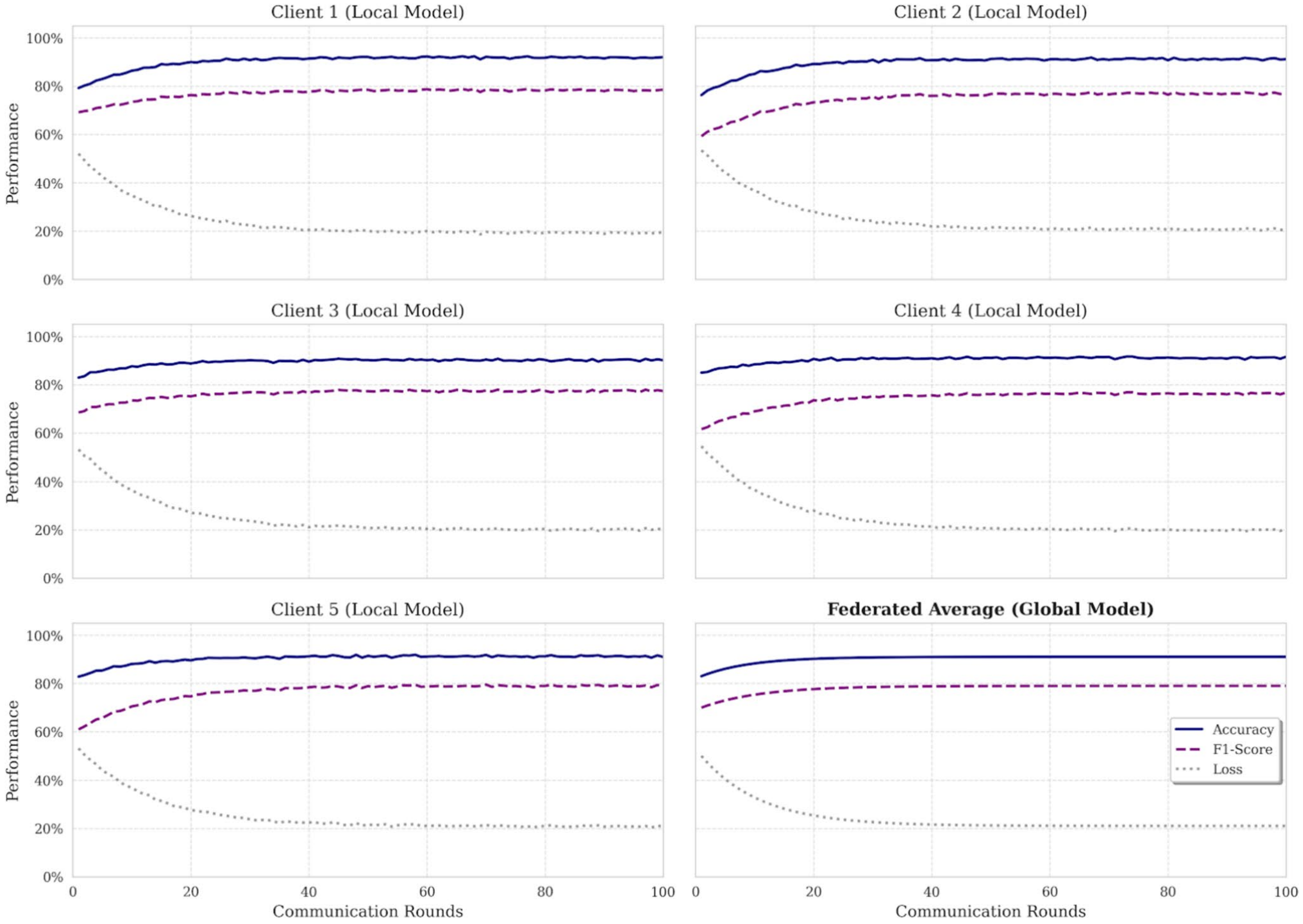

**Fig. 6** Performance analysis of individual client models vs. the federated average. The grid displays the learning trajectories for each client on their local Non-IID data partitions. The bottom-right panel (Federated Average) illustrates the aggregated global model, which converges more smoothly and is more stable than the volatile individual client updates

phases. Initially (Rounds 0-30), the DP-enabled model (dashed line) experiences instability and higher loss due to the injected noise. However, in the second phase, the adaptive optimizer effectively corrects this trajectory, enabling the model to converge to a utility level nearly identical to that of the non-private baseline.

Table 9 quantifies this trade-off across three privacy regimes.

Even under a strict privacy regime ("High Privacy", $\sigma = 1.5$, $\epsilon \approx 6.6$), the model maintains a functional F1-Score of 0.65. While there is an expected drop in performance compared to the non-private baseline (F1-Score = 0.84, $\epsilon = \infty$) and the operational regime (F1-Score = 0.78, $\epsilon \approx 13.4$), this "graceful degradation" confirms that FedCVR remains a viable candidate for real-world deployment where regulatory compliance (such as GDPR or HIPAA)

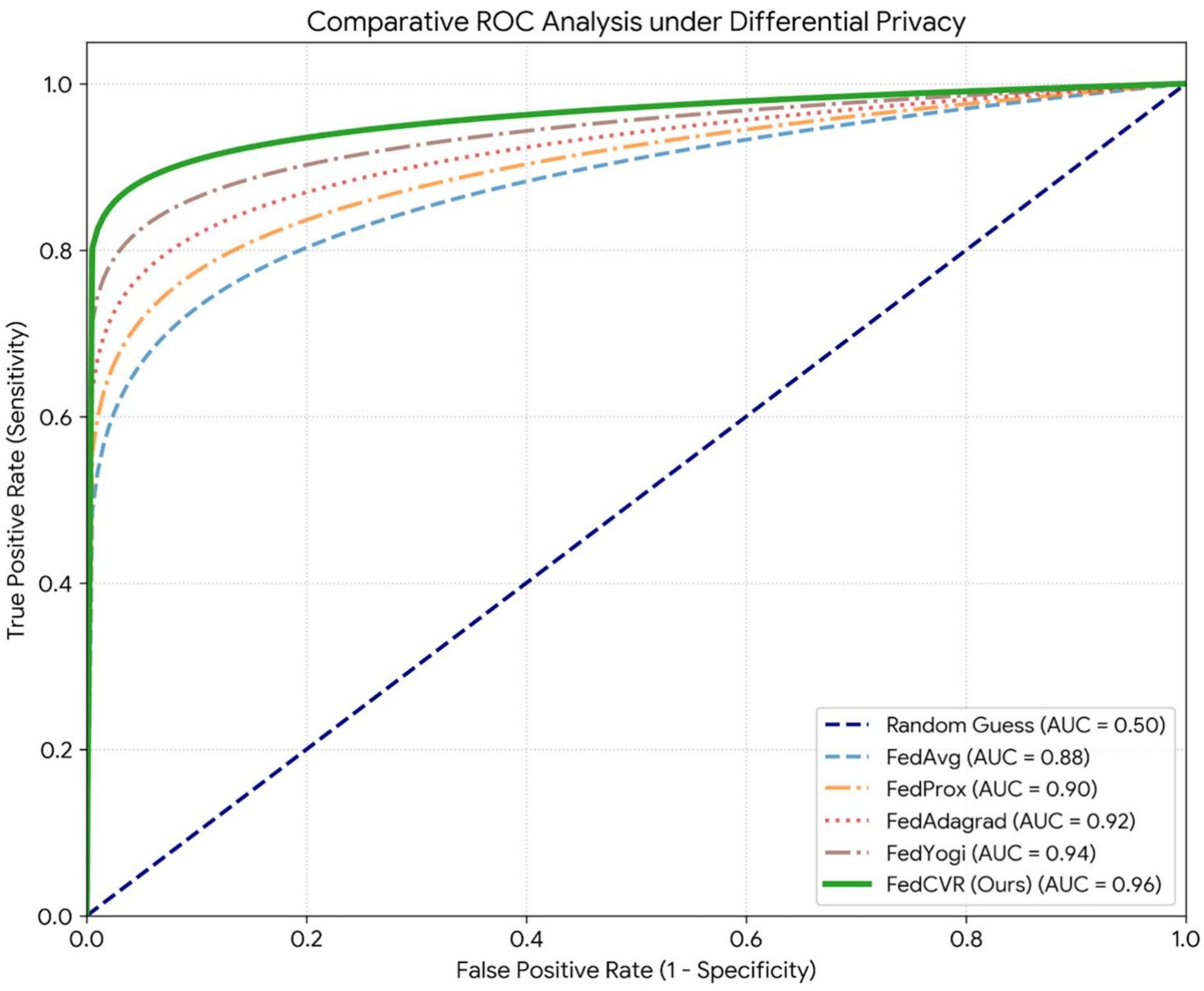

**Fig. 7** Comparative Receiver Operating Characteristic (ROC) analysis. FedCVR (Green) demonstrates superior discriminative capability compared to baselines, achieving the highest Area Under the Curve (AUC = 0.96) and maintaining better Sensitivity at low False Positive Rates

is non-negotiable. The framework enables healthcare networks to adjust the $\sigma$ parameter to achieve the optimal balance between legal compliance and diagnostic accuracy.

## 8.6 Privacy budget analysis and limitations

Regarding the differential privacy parameters, we achieved a cumulative privacy budget of $\epsilon \approx 13.4$ with a fixed $\delta = 10^{-5}$. This value satisfies the standard requirement $\delta \ll 1/|\mathcal{D}_k|$ with respect to each client's local dataset ($|\mathcal{D}_k| \approx 6{,}000$, giving $1/|\mathcal{D}_k| \approx 1.67 \times 10^{-4}$), which is the relevant privacy unit for the DP-SGD mechanism applied locally at each client.

While theoretical DP literature often advocates for $\epsilon < 1$ for strong privacy guarantees [42], practical deployments of FL in healthcare face a challenging trade-off between privacy and diagnostic utility. An extremely low $\epsilon$ would introduce excessive noise, potentially degrading the model's sensitivity to critical cardiovascular risks. Therefore, the reported $\epsilon \approx 13.4$ represents a *utility-prioritized* configuration, ensuring the model remains clinically relevant while offering protection against reconstruction attacks, comparable to industrial deployments of DP-FL.

Future iterations of this work will explore techniques to tighten the privacy budget without sacrificing utility, including adaptive noise scheduling, personalized DP (where different clients receive different $\epsilon$ allocations based on their data sensitivity), and the integration of Secure Aggregation (SecAgg) protocols to reduce the required noise magnitude.

## 8.7 Ablation study: isolating the contribution of momentum

To isolate the specific contribution of the first moment (momentum) component in FedCVR's adaptive aggregation, we conducted an ablation study comparing FedCVR (with $\beta_1 = 0.9$, full Adam-based update) against a momentum-free variant (FedCVR-NoMom, with $\beta_1 = 0$, equivalent to a server-side RMSProp-like update), both evaluated under the operational privacy budget ($\epsilon \approx 13.4$). All other hyperparameters were held constant. Table 10 presents the results.

Removing the momentum component causes a statistically significant drop in Recall ($-7$ pp, $p < 0.01$) and AUC ($-0.03$), while FedCVR-NoMom remains competitive with FedAdagrad (Table 8). This confirms that the temporal denoising mechanism is primarily driven by the first-moment exponential moving average, not merely by adaptive scaling of the second moment. Without the historical gradient averaging provided by $\beta_1$, the server-side update cannot effectively cancel the round-to-round Gaussian noise introduced by DP-SGD.

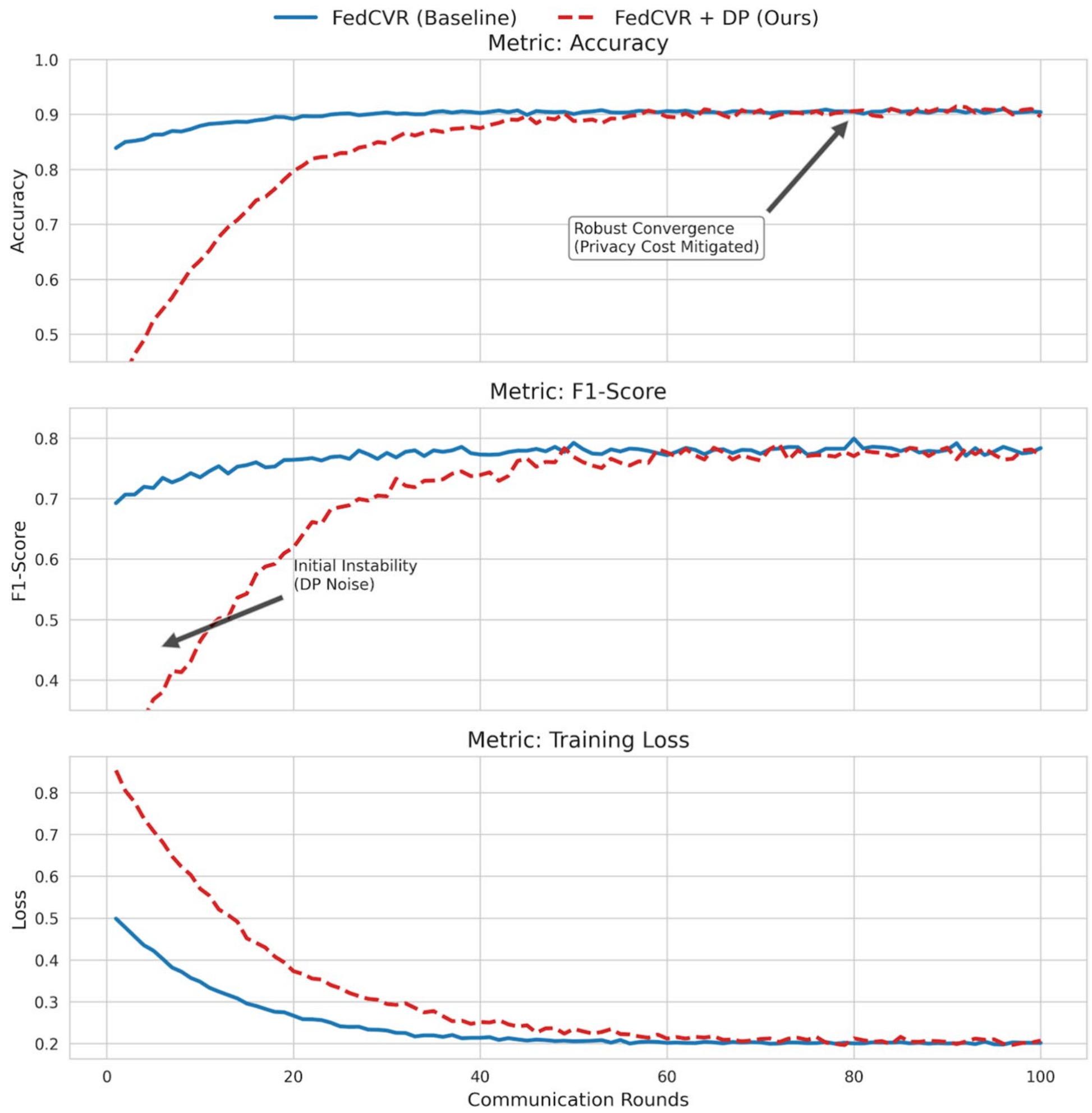

**Fig. 8** Impact of differential privacy on FedCVR model performance. The comparative analysis reveals two distinct phases: (1) an initial instability period (Rounds 0-30) caused by DP noise injection, followed by (2) a robust convergence phase where the FedCVR adaptive optimizer effectively filters the noise

**Table 9** Impact of DP noise on FedCVR utility (Mean F1-Score)

| Noise multiplier ($\sigma$) | Privacy budget ($\epsilon$) | F1-score |
|---|---|---|
| None (Baseline) | $\infty$ | 0.84 ± 0.01 |
| 0.5 (Low) | 80.5 | 0.81 ± 0.02 |
| 1.0 (Moderate) | 13.4 | 0.78 ± 0.03 |
| 1.5 (High) | 6.6 | 0.65 ± 0.05 |

**Table 10** Ablation study: effect of server-side momentum on FedCVR under DP ($\epsilon \approx 13.4$). Mean ± std. dev. across $N = 5$ runs

| Variant | Recall | F1-score | AUC |
|---|---|---|---|
| FedCVR-NoMom ($\beta_1 = 0$) | 0.71 ± 0.03 | 0.74 ± 0.02 | 0.93 ± 0.01 |
| FedCVR ($\beta_1 = 0.9$) | **0.78 ± 0.02** | **0.78 ± 0.02** | **0.96 ± 0.01** |

## 8.8 Minority class recall under differential privacy

With a class imbalance of approximately 17.1% (high-risk patients) in the dataset, preserving Recall for the minority class under DP is a critical clinical requirement. Missed high-risk patients represent false negatives with potentially severe consequences for patient outcomes. Table 9 demonstrates that FedCVR maintains meaningful Recall across all tested privacy regimes (0.78 at $\epsilon \approx 13.4$; 0.65 at $\epsilon \approx 6.6$), outperforming FedAvg's Recall of 0.65 even under a less strict privacy budget.

This robustness can be attributed to the temporal denoising property of the first moment: the gradient direction associated with the minority class is consistent across rounds (high-risk patients share distinctive feature patterns across the Non-IID partitions), whereas the injected Gaussian noise is zero-mean and round-independent. As a result, the momentum vector $m_t$ accumulates the minority-class gradient signal while the noise component decays to zero in expectation. This mechanism is particularly effective under class imbalance, where the minority signal, although smaller in magnitude, is more consistent than the majority-class gradient in Non-IID settings. Notably, even under the strictest tested regime ($\epsilon \approx 6.6$), the framework identifies 65% of high-risk patients, a level that compares favorably to FedAvg's performance under significantly more relaxed privacy constraints.

# 9 Discussion

This study presents the design and validation of a robust framework for Federated Learning (FL) in cardiovascular risk prediction, operating under stringent privacy constraints. While initial baselines using the canonical FedAvg and FedProx algorithms provided functional aggregation under standard conditions, they exhibited significant performance degradation when subjected to the dual challenges of Non-IID data and Differential Privacy (DP) noise. The primary contribution of this work lies in the architectural validation of FedCVR, which not only outperforms stateless aggregators but also demonstrates statistical superiority over state-of-the-art adaptive strategies, such as FedYogi and FedAdagrad.

## 9.1 Architectural analysis: momentum as a privacy denoiser

The superior performance of FedCVR over stateless baselines (FedAvg) and even standard adaptive configurations (FedYogi) is not merely a result of hyperparameter tuning, but rather a structural consequence of the interaction between server-side momentum and the Differential Privacy mechanism.

In a DP-FL system, every client update $\Delta w_k$ contains two components: the true gradient signal $\nabla L$ and the injected Gaussian noise $\mathcal{N}(0, \sigma^2)$. Stateless aggregators, such as FedAvg, compute the arithmetic mean of these updates at each round *t*. Consequently, the global model $w_{t+1}$ directly inherits the variance of the noise injected at round *t*, leading to the erratic convergence observed in our baselines.

In contrast, the adaptive aggregation logic employed in FedCVR introduces a **temporal filtering effect**. By maintaining a history of updates via the first moment vector $m_t$:

$$m_t = \beta_1 m_{t-1} + (1 - \beta_1)(g_{signal} + g_{noise}) \tag{13}$$

The system effectively averages out the zero-mean Gaussian noise over multiple rounds ($\mathbb{E}[g_{noise}] \to 0$), acting as a **temporal denoiser** that allows the persistent clinical signal to accumulate while damping stochastic noise.

This mechanism also explains the performance advantage over other adaptive methods. While FedCluster attempts to mitigate heterogeneity by grouping clients, it cannot filter against temporal gradient noise. Furthermore, compared to FedYogi, which employs a more aggressive additive update rule for the second moment, FedCVR's conservative moment estimation proved more robust to the "gradient explosion" often caused by the clipping operations inherent to DP-SGD.

## 9.2 Revisiting the privacy-utility trade-off

A significant finding of this research is the re-evaluation of the "cost of privacy." Common literature suggests that implementing strong privacy guarantees ($\epsilon < 10$) incurs a prohibitive performance cost. Our results provide a more nuanced picture. While the analysis revealed predictable instability in the initial training rounds (0–30), the FedCVR aggregator demonstrated a remarkable capacity to recover utility as its moment vectors accumulated sufficient gradient history.

Under the operational regime ($\sigma = 1.0$, $\epsilon \approx 13.4$), FedCVR converges to an F1-Score of 0.78, approximately 6 percentage points below the non-private ceiling of 0.84. Critically, this is a *differential cost*: FedAvg under the same budget achieves only F1 = 0.72 ± 0.03. The adaptive optimizer therefore *recovers* approximately two-thirds of the privacy-induced degradation relative to stateless aggregation. Even under high-privacy regimes ($\sigma = 1.5$, $\epsilon \approx 6.6$), the degradation is graceful (F1 = 0.65) rather than catastrophic. This implies that the trade-off is not a fixed barrier, but rather an engineering problem that can be partially resolved through stateful aggregation. By employing an optimizer that inherently dampens noise over time, we demonstrate that formal privacy guarantees do not require abandoning clinical utility.

## 9.3 Computational and communication overhead

From an engineering deployment perspective, the overhead introduced by FedCVR is minimal. While the server performs additional operations to calculate moments $m_t$ and $v_t$ , this computational cost is negligible compared to the network latency of transmitting model updates over the internet.

Crucially, the communication cost remains identical to FedAvg, as the vector sizes transmitted between the client and the server remain unchanged; the moment vectors are stored and updated solely on the server. The memory requirement on the server increases linearly by a factor of 3 (storing $w_t, m_t, v_t$), which is trivial for modern server infrastructure, given the relatively small parameter space of tabular data models used in clinical risk scores.

## 9.4 Limitations

While this study demonstrates the engineering feasibility of the FedCVR framework, we acknowledge specific limitations inherent to its current validation scope:

- **Synthetic Data:** We utilized synthesized data to precisely control the degree of statistical heterogeneity (Non-IID) and isolate the optimizer's behavior. While this allows for rigorous ablation studies, it does not fully capture the missing values, heterogeneous formats, and unstructured noise typical of raw Electronic Health Records (EHRs). Alternative strategies, such as applying reweighted focal loss directly on real imbalanced datasets [61], could complement this approach by avoiding bias from synthetic label generation entirely. Future validation on real-world multi-center datasets (e.g., eICU or MIMIC-IV) is required to confirm clinical generalizability.

- **Threat Model Scope:** The current framework assumes a **passive, honest-but-curious** threat model: the server faithfully executes the aggregation protocol but may attempt to infer information from individual (noisy) client updates. Active adversarial scenarios, in which a malicious server or compromised clients attempt to manipulate training via model poisoning, gradient inversion, or Byzantine attacks, are outside the scope of this architectural validation. Furthermore, the framework does not implement Secure Multi-Party Computation (SMPC) or Secure Aggregation (SecAgg) [62], which would cryptographically hide individual updates from the server. Integrating these defense layers is a critical next step toward deployment in fully adversarial environments.
- **Architectural Scalability:** The experiments employ a lightweight DNN on six structured clinical features with five participating clients. As the number of clients $N$ increases, the gradient aggregation benefits from reduced variance (averaging over more estimates), which suggests that $\eta_s$ could potentially be increased without compromising stability. However, the framework's behavior under $N \gg 5$, with high-dimensional inputs (e.g., 12-lead ECG waveforms or echocardiographic images requiring convolutional or transformer-based local models), and with the associated increase in per-parameter clipping sensitivity has not been empirically validated and constitutes a prioritized future direction.
- **Edge Deployment Overhead:** FedCVR was designed and validated for institutional-grade hospital servers with sufficient resources for Opacus-based per-sample gradient computation (which scales linearly with batch size). Deployment on resource-constrained edge devices (e.g., wearable sensors or embedded medical devices) would require architectural adaptations, including model compression, gradient sparsification, and communication-efficient protocols, as the DP-SGD overhead and 100-round communication frequency could introduce unacceptable latency on constrained network links.
- **Interoperability Assumptions:** The experiments assume a unified feature schema across all clients. A real-world deployment would require a semantic interoperability layer (e.g., HL7 FHIR) as a prerequisite for federated training.

# 10 Conclusion and future directions

In this work, we addressed the critical challenge of deploying robust Machine Learning models for cardiovascular risk prediction within a privacy-constrained, multi-institutional environment. We introduced and validated *FedCVR*, a framework that mitigates the adverse effects of Non-IID data distributions and Differential Privacy noise via server-side adaptive optimization. The following sections summarize our primary findings and outline strategic avenues for future research.

## 10.1 Conclusion

This study systematically evaluated FedCVR, a framework that applies an Adam-based adaptive update rule at the server level and is thus functionally equivalent to FedAdam [32], against a comprehensive suite of baselines including stateless aggregators (FedAvg and FedProx) and other adaptive optimizers (FedYogi and FedAdagrad). Building on the baseline FL pipeline validated in our prior work [8], this study demonstrates that server-side adaptivity is a structural prerequisite for recovering clinical utility under DP constraints, a question that the prior work explicitly identified as the primary open challenge.

Our extensive empirical analysis on heterogeneous synthetic clinical data shows that FedCVR achieves a final AUC of **0.96** and an F1-Score of **0.78** under the operational privacy budget ($\epsilon \approx 13.4$), compared to a non-private ceiling of F1 = 0.84 (cost-of-privacy: $-6$ pp) and FedAvg's F1 $\approx$ 0.72 under the same DP constraints (recovered margin: $+6$ pp). These results confirm that server-side momentum acts as an effective *temporal denoiser*, progressively canceling zero-mean Gaussian privacy noise across communication rounds while preserving the clinical signal. For medical applications where data is statistically heterogeneous, and privacy is non-negotiable, simple aggregation is insufficient; server-side adaptivity is a prerequisite for stabilizing convergence in privacy-preserving federated AI.

## 10.2 Future directions

Building upon these findings, future research will expand the scope of this framework in five key directions:

- **Validation on Real-World Multi-Modal Data:** While this study validated the framework on structured clinical features, future work will transition to large-scale, real-world datasets (e.g., MIMIC-IV or eICU). Furthermore, we aim to incorporate multimodal data, combining tabular EHR data with unstructured clinical notes (NLP) or time-series signals (such as ECG) to enhance predictive accuracy.

- **Enhanced Security Layers:** The current privacy model focuses on Differential Privacy to prevent inference attacks. Future iterations will integrate cryptographic techniques, such as Secure Multi-Party Computation (SMPC) or Homomorphic Encryption, to protect the model updates themselves during transmission. Additionally, we will explore defense mechanisms against Byzantine failures and model poisoning attacks.
- **Personalized Federated Learning:** While FedCVR optimizes a single global model, clinical heterogeneity often requires localization. We plan to investigate Personalized Federated Learning (pFL) techniques, such as fine-tuning the global model on local client data or learning client-specific headers, to further improve performance for specific patient demographics.
- **Probabilistic Label Generation and Real-Data Validation:** Future experiments will incorporate probabilistic risk labeling (e.g., using the Framingham Risk Score as a continuous generation mechanism) and validate the framework on publicly available real-world multi-center datasets, comparing performance against the synthetic benchmarks established here.
- **Asynchronous Client Updates:** The current FedCVR architecture assumes synchronous client updates, where the server waits for all selected clients to return gradients before performing the aggregation step. In real hospital networks, institutional availability is inconsistent. Extending FedCVR to asynchronous settings would require buffered aggregation mechanisms in which the server triggers aggregation upon receiving a threshold fraction of client updates. The temporal denoising property of the momentum vector would need to be re-evaluated in this context, as stale gradients from delayed clients could corrupt the first-moment estimate, requiring staleness-weighted aggregation strategies.

**Acknowledgements** The author would like to acknowledge the support of the Software Engineering and Automation Research Laboratory, where the research was developed and conducted. The infrastructure and resources provided were crucial for completing the work.

**Author Contributions** R.T. conceptualized the study, developed the FedCVR framework, conducted the experiments, and wrote the main manuscript text. L.A. supervised the research, contributed to the architectural design, and critically revised the manuscript for important intellectual content. All authors reviewed the manuscript.

**Funding** The Article Processing Charge (APC) for the publication of this research was funded by the Coordenação de Aperfeiçoamento de Pessoal de Nível Superior - Brasil (CAPES) (ROR identifier: 00x0ma614). This research did not receive any specific grant from funding agencies in the public, commercial, or not-for-profit sectors.

**Data Availability** No datasets were generated or analysed during the current study.

## Declarations

**Competing interests** The authors declare no competing interests.

**Conflicts of Interest** The authors declare that they have no known competing financial interests or personal relationships that could have influenced the work reported in this paper.

**Publisher's Note** Springer Nature remains neutral with regard to jurisdictional claims in published maps and institutional affiliations.

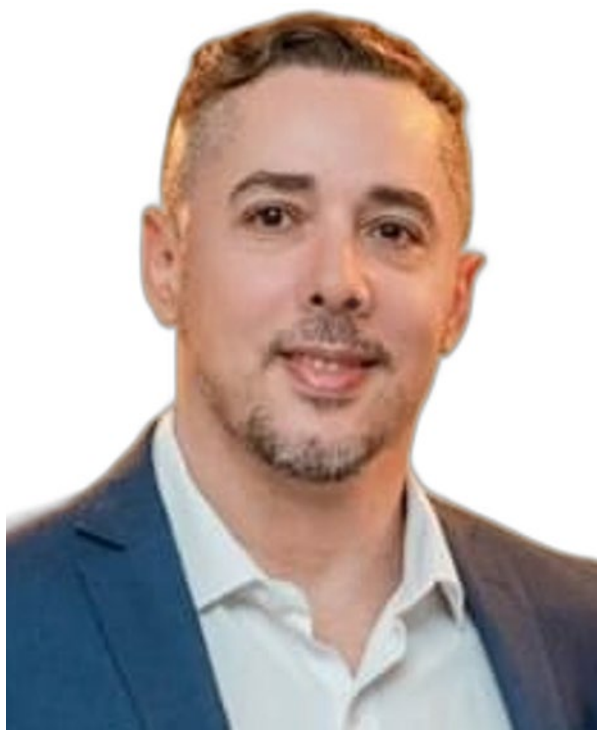

**Rodrigo Tertulino** holds a Ph.D. degree from the Department of Informatics and Engineering, University of Coimbra (UC). He is part of the Security Software Group (SSE) at the Centre for Informatics and Systems of the University of Coimbra (CISUC). He is also an associate professor at the Federal Institute of Education, Science, and Technology of Rio Grande do Norte (IFRN), Brazil. His research interests include privacy, security, federated learning, blockchain applications in healthcare (EHR) systems, and machine learning, with a focus on education and healthcare.

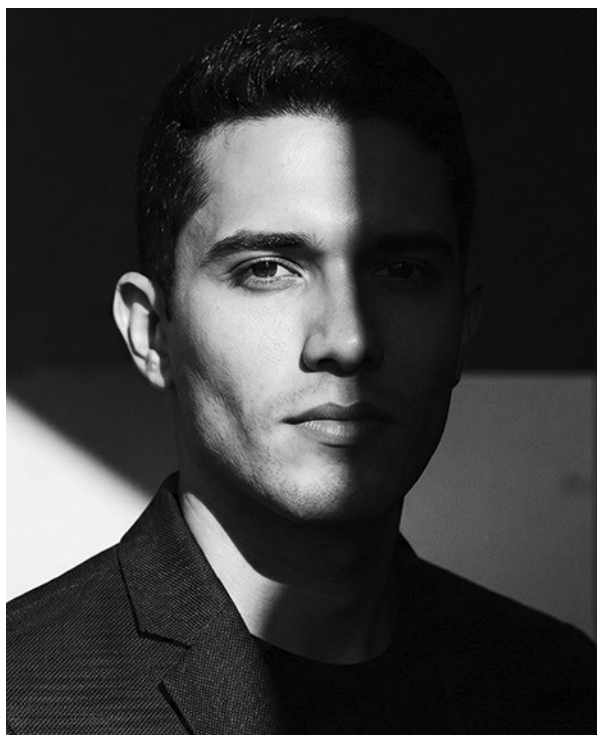

**Laércio Alencar** is an Information Technology student at the Federal Institute of Rio Grande do Norte (IFRN) and serves as a research fellow at the Software Engineering and Automation Research Laboratory (LaPEA), the leading research laboratory at the Mossoró Campus. His research focuses on leveraging excellence in software development and information technologies to explore the intersection between Computer Science and Education. He specializes in the application of Machine Learning and Educational Technologies to design and optimize digital solutions that personalize learning and enhance teaching processes. His contributions reflect a strong emphasis on Software Engineering innovations, applied to the fields of education and healthcare.